\documentclass[12pt]{article}

\PassOptionsToPackage{numbers, compress}{natbib}

\usepackage[top=3cm, bottom=3cm, left=3cm, right=3cm]{geometry}
\usepackage{here}

\usepackage[utf8]{inputenc} 
\usepackage[T1]{fontenc}     
\usepackage{hyperref}          
\usepackage{url}                   
\usepackage{booktabs}        
\usepackage{amsfonts}        
\usepackage{nicefrac}          
\usepackage{microtype}       

\usepackage[dvipdfmx]{graphicx}

\usepackage{amsmath}
\usepackage{listings}
\usepackage{color}
\usepackage{upquote}

\newcommand{\diag}{{\rm diag}}
\newcommand{\tr}{{\rm tr}}
\newcommand{\E}{{\rm E}}

\newcommand{\Cov}{{\rm Cov}}
\newcommand{\eig}{{\rm eig}}
\newcommand{\sign}{{\rm sign}}

\begin{document}
\noindent\textbf{\Large On the achievability of blind source separation for high-dimensional nonlinear source mixtures} \\
\\
\noindent\textbf{Takuya Isomura} \\
Laboratory for Neural Computation and Adaptation, RIKEN Center for Brain Science, Wako, Saitama 351-0198, Japan \\
Brain Intelligence Theory Unit, RIKEN Center for Brain Science, Wako, Saitama 351-0198, Japan \\
\texttt{takuya.isomura@riken.jp} \\
\\
\noindent\textbf{Taro Toyoizumi} \\
Laboratory for Neural Computation and Adaptation, RIKEN Center for Brain Science, Wako, Saitama 351-0198, Japan \\
Department of Mathematical Informatics, Graduate School of Information Science and Technology, The University of Tokyo, Bunkyo-ku, Tokyo 113-8656, Japan \\
\texttt{taro.toyoizumi@riken.jp}
\newpage

\section*{Abstract}
For many years, a combination of principal component analysis (PCA) and independent component analysis (ICA) has been used for blind source separation (BSS). However, it remains unclear why these linear methods work well with real-world data that involve nonlinear source mixtures. This work theoretically validates that a cascade of linear PCA and ICA can solve a nonlinear BSS problem accurately---when the sensory inputs are generated from hidden sources via nonlinear mappings with sufficient dimensionality. Our proposed theorem, termed the asymptotic linearization theorem, theoretically guarantees that applying linear PCA to the inputs can reliably extract a subspace spanned by the linear projections from every hidden source as the major components---and thus projecting the inputs onto their major eigenspace can effectively recover a linear transformation of the hidden sources. Then, subsequent application of linear ICA can separate all the true independent hidden sources accurately. Zero-element-wise-error nonlinear BSS is asymptotically attained when the source dimensionality is large and the input dimensionality is sufficiently larger than the source dimensionality. Our proposed theorem is validated analytically and numerically. Moreover, the same computation can be performed by using Hebbian-like plasticity rules, implying the biological plausibility of this nonlinear BSS strategy. Our results highlight the utility of linear PCA and ICA for accurately and reliably recovering nonlinearly mixed sources---and further suggest the importance of employing sensors with sufficient dimensionality to identify true hidden sources of real-world data.


\section{Introduction}
Blind source separation (BSS) involves the separation of mixed sensory inputs into their hidden sources without knowledge of the manner in which they were mixed (Cichocki et al., 2009; Comon \& Jutten, 2010). Among the numerous BSS methods, a combination of principal component analysis (PCA) (Pearson, 1901; Oja, 1982; Oja, 1989; Sanger, 1989; Xu, 1993; Jolliffe, 2002) and independent component analysis (ICA) (Comon, 1994; Bell \& Sejnowski, 1995; Bell \& Sejnowski, 1997; Amari et al., 1996; Hyvarinen \& Oja, 1997) is one of the most widely used approaches. In this combined PCA--ICA approach, PCA yields a low-dimensional concise representation, i.e., the major principal components, of sensory inputs that most suitably describes the original high-dimensional redundant inputs. Whereas, ICA provides a representation, i.e., encoders, that separates the compressed sensory inputs into independent hidden sources. A classical setup for BSS assumes a linear generative process (Bell \& Sejnowski, 1995), in which sensory inputs are generated as a linear superposition of independent hidden sources. The linear BSS problem has been extensively studied both analytically and numerically (Amari et al., 1997; Oja \& Yuan, 2006; Erdogan, 2009), where the cascade of PCA and ICA is guaranteed to provide the optimal linear encoder that can separate sensory inputs into their true hidden sources, up to permutations and sign-flips (Baldi \& Hornik, 1989; Chen et al., 1998; Papadias, 2000; Erdogan, 2007).

Another most crucial perspective is applicability of BSS methods to real-world data generated from a nonlinear generative process. In particular, the aim of nonlinear BSS is to identify the inverse of a nonlinear generative process that generates sensory inputs and thereby infer their true independent hidden sources based exclusively on the sensory inputs. Although the cascade of linear PCA and ICA has been applied empirically to real-world BSS problems (Calhoun et al., 2009), no one has yet theoretically proven that this linear BSS approach can solve a nonlinear BSS problem. To address this gap, this work demonstrates mathematically that the cascade of PCA and ICA can solve a nonlinear BSS problem accurately when the source dimensionality is large and the input dimensionality is sufficiently larger than the source dimensionality so that various nonlinear mappings from sources to inputs can be determined from the inputs themselves.

In general, there are five requirements for solving the nonlinear BSS problem. The first two requirements are related to the representation capacity of the encoder: (1) the encoder's parameter space must be sufficiently large to accommodate the actual solution that can express the inverse of the true generative process; (2) however, this parameter space should not be too large; otherwise, a nonlinear BSS problem can have infinitely many spurious solutions wherein all encoders are independent but dissimilar to the true hidden sources (Hyvarinen \& Pajunen, 1999; Jutten \& Karhunen, 2004). Hence, it is important to constrain the representation capacity of the encoder in order to satisfy these opposing requirements. A typical approach for solving the nonlinear BSS problem involves using a multilayer neural network---with nonlinear activation functions---that learns the inverse of the generative process (Lappalainen \& Honkela, 2000; Karhunen, 2001; Hinton \& Salakhutdinov, 2006; Kingma \& Welling, 2013; Dinh et al., 2014). The remaining three requirements are related to the unsupervised learning algorithms used to identify the optimal parameters for the encoder: (3) the learning algorithm must have a fixed point at which the network expresses the inverse of the generative process; (4) the fixed point must be linearly stable so that the learning process converges to the solution; and (5) the probability of not converging to this solution should be small; i.e., most realistic initial conditions must be within the basin of attraction of the true solution.

Approaches using a nonlinear multilayer neural network satisfy Requirement 1 when the number of neurons in each layer is sufficient (c.f., universality (Cybenko, 1989; Hornik et al., 1989; Barron, 1993)); moreover, learning algorithms that satisfy Requirements 3 and 4 are also known (Dayan et al., 1995; Friston, 2008; Friston et al., 2008). However, reliable identification of the true hidden sources is still necessary, because the encoder can have infinitely many spurious solutions if its representation capacity is too large (i.e., if Requirement 2 is violated). As previously indicated, to the best of our knowledge, there is no theoretical proof that confirms a solution for a nonlinear BSS problem (Hyvarinen \& Pajunen, 1999; Jutten \& Karhunen, 2004), except for some cases wherein temporal information---such that each independent source has its own dynamics---is available (Hyvarinen \& Morioka, 2016; Hyvarinen \& Morioka, 2017; Khemakhem et al., 2020). Moreover, even when Requirement 2 is satisfied, there is no guarantee that a learning algorithm will converge to the true hidden source representation, because it might be trapped in a local minimum wherein outputs are still not independent of each other. Thus, for a nonlinear BSS problem, it is of paramount importance to simplify the parameter space of the inverse model in order to remove spurious solutions and prevent the learning algorithm from attaining a local minimum (to satisfy Requirements 2 and 5) while retaining its capacity to represent the actual solution (Requirement 1). Hence, in this work, we apply a linear approach to solving a nonlinear BSS problem in order to ensure Requirements 2 and 5 are satisfied. We demonstrate that the cascade of PCA and ICA can reliably identify a good approximation of the inverse of nonlinear generative processes asymptotically under the condition where the source dimensionality is large and the input dimensionality is sufficiently larger than the source dimensionality (thus satisfying Requirements 1--5). Although such a condition is different from the case typically considered by earlier works---where the sources and inputs have the same dimensionality---the condition we consider turns out to be apt for the mathematical justification of the achievability of the nonlinear BSS.


\section{Results}
\subsection{Overview}
Our proposed theorem, referred to as asymptotic linearization theorem, is based on an intuition that when the dimensionality of sensory inputs is significantly larger than that of hidden sources, these inputs must involve various linear and nonlinear mappings of all hidden sources---thus providing sufficient information to identify the true hidden sources without ambiguity using an unsupervised learning approach. We consider that $N_s$-dimensional hidden sources $s \equiv (s_1, \dots, s_{N_s})^T$ generate $N_x$-dimensional sensory inputs $x \equiv (x_1, \dots, x_{N_x})^T$ through a generative process characterized by an arbitrary nonlinear mapping $x \equiv F(s)$ (Fig. 1). Here, the hidden sources are supposed to follow independently an identical probability distribution with zero mean and unit variance $p_s(s) \equiv \prod_i p_i(s_i)$. A class of nonlinear mappings $F(s)$ can be universally approximated by a specific but generic form of two-layer network. Suppose $A \in \mathbb{R}^{N_f \times N_s}$ and $B \in \mathbb{R}^{N_x \times N_f}$ as higher- and lower-layer mixing matrices, respectively, $a \in \mathbb{R}^{N_f}$ as a constant vector of offsets, $y \equiv As$ as a linear mixture of sources, $f(\bullet): \mathbb{R} \mapsto \mathbb{R}$ as a nonlinear function, and $f \equiv (f_1, \dots, f_{N_f})^T \equiv f( A s + a ) \equiv f( y + a )$ as $N_f$-dimensional nonlinear bases. The sensory inputs are given as
\begin{equation}
\label{x}
x = B f( A s + a ),
\end{equation}
or equivalently $x = B f( y + a ) = B f$. This expression using the two-layer network is universal in the component-wise sense \footnote{In this work, based on the literature (Cybenko, 1989), we define the universality in the component-wise sense as the condition wherein each element of $x = F(s)$ is approximated using the two-layer network with an approximation error smaller than an arbitrary $\delta > 0$, i.e., $\mathrm{sup}_s[|F_j(s) - B_j f(As+a)|] < \delta$ for $j = 1,\dots,N_x$ when $N_f$ is sufficiently large. Note that $F_j(s)$ is the $j$-th element of $F(s)$ and $B_j$ is the $j$-th row vector of $B$. It is known that when $f$ is a sigmoidal function and parameters are selected appropriately, the approximation error can be upper-bounded by the order $N_f^{-1}$ (Barron, 1993); hence, $\E[|F_j(s) - B_j f(As+a)|^2] \leq \mathcal{O}(N_f^{-1})$. This relationship holds true when $N_f \gg N_s \gg 1$, irrespective of $N_x$.} (Cybenko, 1989; Hornik et al., 1989; Barron, 1993) and each element of $x$ can represent an arbitrary mapping $x=F(s)$ as $N_f$ increases by adjusting parameters $a$, $A$, and $B$. We further suppose that each element of $A$ and $a$ is independently generated from a Gaussian distribution $\mathcal{N}[0,1/N_s]$, which retains its universality (Rahimi \& Recht, 2008a; Rahimi \& Recht, 2008b) as long as $B$ is tuned to minimize the mean squared error $\E[ |B f - F(s)|^2 ]$. Here, $\E[\bullet]$ describes the average over $p_s(s)$. The scaling of $A$ is to ensure that the argument of $f$ is of order 1. The $N_s^{-1/2}$ order offset $a$ is introduced to this model to express any generative process $F(s)$; however, it is negligibly small relative to $As$ for large $N_s$. The whole system, including generative process and neural network, is depicted in Fig. 1 (left). The corresponding equations are summarized in Fig. 1 (right).


\begin{figure}[H]
  \centering
  \includegraphics[width=1.00\linewidth]{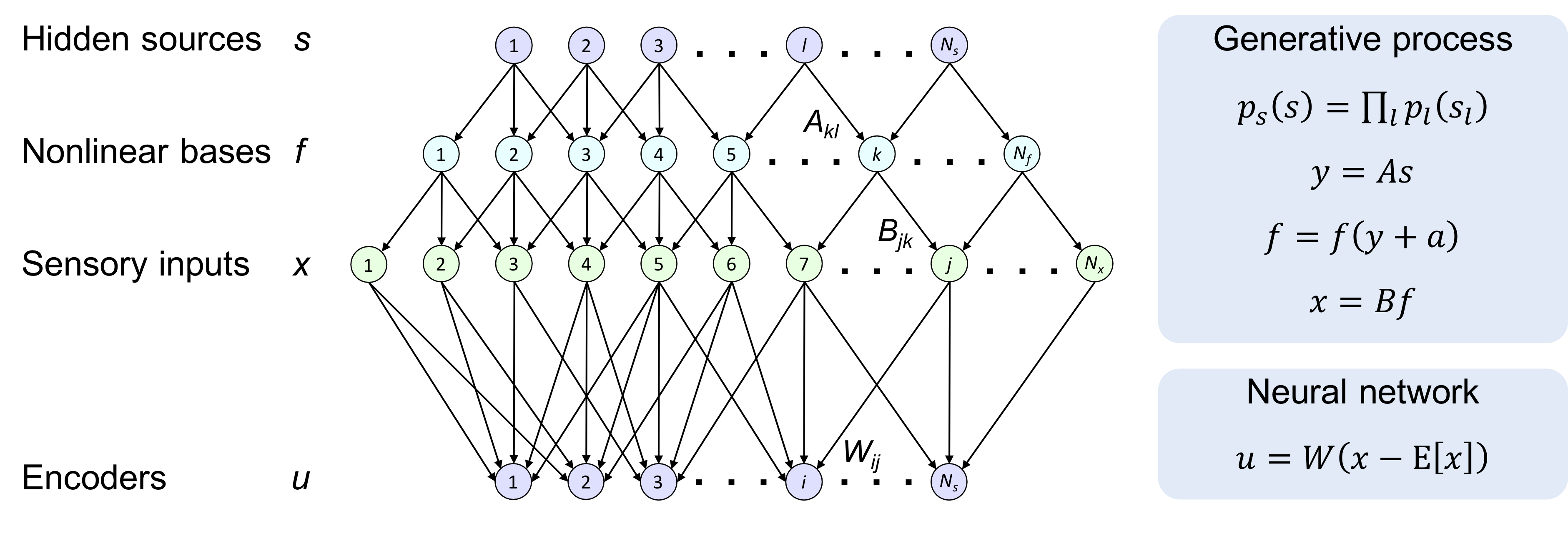}
\end{figure}
\noindent Figure 1. Structures of nonlinear generative process (top) and linear neural network (bottom). Hidden sources $s = (s_1, \dots, s_{N_s})^T$ generate sensory inputs $x = (x_1, \dots, x_{N_x})^T$ via nonlinear mappings characterized by nonlinear bases $f = (f_1, \dots, f_{N_f})^T$, where $A = \{A_{kl}\}$ and $a = (a_1, \dots, a_{N_f})^T$ are Gaussian distributed higher-layer mixing weights and offsets, respectively, and $B = \{B_{jk}\}$ is a lower-layer mixing weight matrix. Encoders comprise a single-layer linear neural network, where $u = (u_1, \dots, u_{N_s})^T$ are neural outputs and $W = \{W_{ij}\}$ is a synaptic weight matrix. Equations in the right-hand-side panels summarize the generation of sensory inputs from hidden sources through nonlinear mixtures, and the inversion of this process using a linear neural network. \\


For analytical tractability, we decompose nonlinear bases $f$ into the sum of linear and nonlinear parts of hidden sources as follows: {\it linear components} of the hidden sources in the bases are defined as $H s$ using a coefficient matrix $H$ that minimizes the mean squared error, $H \equiv \arg\min_H \E[|f - \E[f] - H s|^2]$. Such a coefficient matrix is computed as $H = \E[f s^T]$. The remaining part $\phi \equiv f - \E[f] - H s$ is referred to as {\it nonlinear components} of the hidden sources, which are orthogonal (uncorrelated) to $s$ (i.e., $\E[\phi s^T]=O$). This definition of linear and nonlinear components is unique. Thus, the sensory inputs (equation \eqref{x}) are decomposed into linear and nonlinear transforms of the hidden sources,
\begin{equation}
\label{x2}
x - \E[x] = \underbrace{BHs}_{\mathrm{signal}} + \underbrace{B\phi}_{\mathrm{residual}}.
\end{equation}
The first term on the right-hand side represents the {\it signal} comprising the linear components of the hidden sources, whereas the second term represents the {\it residual} introduced via the nonlinearity in the generative process. Further, because $\phi$ is uncorrelated with $s$, the covariance of the bases is decomposed into their linear and nonlinear counterparts $\Cov[f] \equiv \E[(f-\E[f])(f-\E[f])^T] = \Cov[Hs] + \Cov[\phi] = HH^T + \Sigma$, where $\Sigma \equiv \Cov[\phi]$ indicates the covariance of $\phi$. Thus, the covariance of the sensory inputs $\Cov[x] \equiv B\Cov[f]B^T$ can be decomposed into the signal and residual covariances,
\begin{equation}
\label{Cov_x}
\Cov[x] = \underbrace{BHH^TB^T}_{\mathrm{signal\;covariance}} + \underbrace{B\Sigma B^T}_{\mathrm{residual\;covariance}}.
\end{equation}
Crucially, the signal covariance has only $N_s$ nonzero eigenvalues when $N_f > N_s$ owing to low-column-rank matrix $H$, whereas the eigenvalues of the residual covariance are distributed in the $N_f$-dimensional eigenspace. This implies that if the norms of the linear and nonlinear components in the inputs are in a similar order and $N_s$ is sufficiently large, the eigenvalues of the signal covariance (i.e., linear components) are $N_f/N_s$ order times larger than those of the residual covariance (i.e., nonlinear components). We will prove in the following sections that, when $N_s \gg 1$, this property derives from the fact that elements of $A$ are Gaussian distributed and singular values of $B$ are of order 1.

In what follows, we demonstrate that owing to the aforementioned property, the first $N_s$ major principal components of the input covariance precisely match the signal covariance when the source dimensionality is large and the input dimensionality is sufficiently larger than the source dimensionality. Consequently, projecting the inputs onto the subspace spanned by major eigenvectors can effectively extract the linear components in the inputs. Moreover, the same projection can effectively filter out the nonlinear components in the inputs, because the majority of the nonlinear components are perpendicular to the major eigenspace. Thus, applying PCA to the inputs enables the recovery of a linear transformation of all the true hidden sources of the mixed sensory inputs with a small estimation error. This property, termed asymptotic linearization, enables the reduction of the original nonlinear BSS problem to a simple linear BSS problem, consequently satisfying Requirements 1--5. In the remainder of this paper, we mathematically validate this theorem for a wide range of nonlinear setups, and thereby demonstrate that linear encoder neural networks can perform the nonlinear BSS in a self-organizing manner \footnote{In this paper, we refer to the neural network as the encoder, because networks that convert the input data into a different (lower-dimensional) code or representation are widely recognized as encoders in the literature on machine learning (Hinton \& Salakhutdinov, 2006; Goodfellow, Bengio \& Courville, 2016). However, one may think that the generative process encodes the hidden sources into the sensory inputs through nonlinear mixtures. From this viewpoint, one may term the neural network as the decoder that decodes the inputs to recover the original sources.}. We assume that $N_x = N_f \gg N_s \gg 1$ throughout the manuscript \footnote{Our numerical simulations suggest that the system behaves similarly for $N_x \geq N_f \geq N_s > 1$ in some cases, although our theorem holds mathematically only when $N_x = N_f \gg N_s \gg 1$.}.


\subsection{PCA can extract linear projections of all true hidden sources}
In this section, we first demonstrate that the major principal components of the input covariance (equation\eqref{Cov_x}) precisely match the signal covariance when $N_f \gg N_s \gg 1$, by analytically calculating eigenvalues of $HH^T$ and $\Sigma$ for the aforementioned system. For analytical tractability, we assume that $f(\bullet)$ is an odd nonlinear function. This assumption does not weaken our proposed theorem because the presumed generative process in equation \eqref{x} remains universal. We further assume that $p_s(s)$ is a symmetric distribution.

Each element of---and any pair of two elements in---a vector $y \equiv As$ is approximately Gaussian distributed for large $N_s$ due to the central limit theorem. The deviation of their marginal distribution $p(y_i,y_j)$ from the corresponding zero-mean Gaussian distribution $p_\mathcal{N}(y_i,y_j) \equiv \mathcal{N}[ 0, \tilde{A}\tilde{A}^T ]$ is order $N_s^{-1}$ because the source distribution is symmetric, where $\tilde{A} \in \mathbb{R}^{2 \times N_s}$ indicates a sub-matrix of $A$ comprising its $i$-th and $j$-th rows (Lemma 1, see Methods for the proof). This asymptotic property allows us to compute $H$ and $\Sigma$ based on the expectation over a tractable Gaussian distribution $p_\mathcal{N}(y_i,y_j)$---as a proxy for the expectation over $p(y_i,y_j)$---as the leading order for large $N_s$, despite the fact that $s$ actually follows a non-Gaussian distribution (please ensure that $p(y_i,y_j)$ converges to $p_\mathcal{N}(y_i,y_j)$ in the large $N_s$ limit). Let us denote the expectation over $p_\mathcal{N}(y_i,y_j)$ as $\E_\mathcal{N}[\bullet] \equiv \int \bullet p_\mathcal{N}(y_i,y_j) dy_idy_j$ to distinguish it from $\E[\bullet]$. The latter can be rewritten as $\E[\bullet] = \E_\mathcal{N}[\bullet (1 + G(y_i,y_j)/N_s)]$ using an order-one function $G(y_i,y_j)$ that characterizes the deviation of $p(y_i,y_j)$ from $p_\mathcal{N}(y_i,y_j)$; thus, the deviation caused by this approximation is negligibly smaller than the leading order in the following analyses.

Owing to this asymptotic property, the coefficient matrix $H$ is computed as follows: the pseudo inverse of $A$, $A^+ \equiv (A^TA)^{-1}A^T$, satisfies $A^+ A = I$ and $A^{+T} = A^{T+}$. Thus, we have $H = HA^TA^{+T} = \E[fy^T]A^{+T}$. It can be approximated by $\E_\mathcal{N}[fy^T]A^{+T}$ as the leading order from Lemma 1. From the integration by parts, we obtain $\E_\mathcal{N}[fy^T] = -\int f p'_\mathcal{N}(y)^T dy \; AA^T = \int \diag[f'] p_\mathcal{N}(y) dy \; AA^T = \diag[\E_\mathcal{N}[f']]AA^T$. This relationship is also known as the Bussgang theorem (Bussgang, 1952). Thus, $\E_\mathcal{N}[fy^T]A^{+T} = \diag[\E_\mathcal{N}[f']]A$ is order $N_s^{-1/2}$ for a generic odd function $f(\bullet)$ (as it is the product of order 1 diagonal matrix $\diag[\E_\mathcal{N}[f']]$ and order $N_s^{-1/2}$ matrix $A$). According to Lemma 1, the difference between $H$ and $\diag[\E_\mathcal{N}[f']]A$ is order $N_s^{-3/2}$ as it is $N_s^{-1}$ order times smaller than the leading order (see Remark in Section 4.1 for details). Hence, we obtain
\begin{equation}
\label{H}
H = \diag[\E_\mathcal{N}[f']]A + \mathcal{O}\left(N_s^{-3/2}\right) = \overline{f'}A + \mathcal{O}\left(N_s^{-1}\right).
\end{equation}
In the last equality, the scalar coefficient is expressed as $\overline{f'} \equiv \int f'(\xi)\exp(-\xi^2/2){\rm d}\xi/\sqrt{2\pi}$ using the expectation over a unit Gaussian variable $\xi$. The order $N_s^{-1}$ error term includes the non-Gaussian contributions of $y$ and the effect of the order $N_s^{-1/2}$ deviation of $\diag[\E_\mathcal{N}[f']]$ from $\overline{f'}$, where the latter yields the order $N_s^{-1}$ error owing to the product of order $N_s^{-1/2}$ diagonal matrix $(\diag[\E_\mathcal{N}[f']] - \overline{f'}I)$ and order $N_s^{-1/2}$ matrix $A$. Due to low column-rank of matrix $A$, $AA^T$ has $N_s$ nonzero eigenvalues, all of which are $N_f/N_s$ as the leading order, because elements of $A$ are independent and identically distributed variables sampled from a Gaussian distribution $\mathcal{N}[0,1/N_s]$ (Marchenko \& Pastur, 1967). The same scaling of eigenvalues also holds for $HH^T$ because the difference between the nonzero singular values of $H$ and those of $\overline{f'}A$---caused by the order $N_s^{-1/2}$ deviation of $\diag[\E_\mathcal{N}[f']]$ from $\overline{f'}I$---is smaller than order $(N_f/N_s)^{1/2}$.

Next, we characterize the covariance matrix $\Sigma = \Cov[\phi]$ using the aforementioned asymptotic property. The nonlinear components can be cast as a function of $y$, $\phi(y) = f(y+a) - \E[f] - \E[fy^T] A^{+T} A^+ y$. For large $N_s$, the covariance of $y$, $\Cov[y]=AA^T$, is close to the identity matrix---and the deviation $(AA^T-I)$ is order $N_s^{-1/2}$ for each element---because elements of $A$ are Gaussian distributed. This weak correlation between $y_i$ and $y_j$ ($i\ne j$) leads to a weak correlation of their functions $\phi_i$ and $\phi_j$. Thus, by computing the Taylor expansion of $\Cov_\mathcal{N}[\phi_i, \phi_j] \equiv \E_\mathcal{N}[\phi_i \phi_j] - \E_\mathcal{N}[\phi_i]\E_\mathcal{N}[\phi_j]$ with respect to the small covariance $\Cov_\mathcal{N}[y_i, y_j] \equiv \E_\mathcal{N}[y_i y_j]$, we obtain (Toyoizumi \& Abbott, 2011) (Lemma 2 in Methods):
\begin{equation}
\label{Cov_phi}
\Cov_\mathcal{N}[\phi_i, \phi_j] = \sum_{n=1}^\infty \frac{\E_\mathcal{N}\left[\phi_i^{(n)}\right] \E_\mathcal{N}\left[\phi_j^{(n)}\right]}{n!} \Cov_\mathcal{N}[y_i, y_j]^n
\end{equation}
for $i \neq j$. Here, $\Cov_\mathcal{N}[y_i, y_j]^n$ is order $N_s^{-n/2}$. Because $\E_\mathcal{N}[\phi^{(1)}_i] = \mathcal{O}(N_s^{-1})$, $\E_\mathcal{N}[\phi^{(2)}_i] = \mathcal{O}(N_s^{-1/2})$, and $|\E_\mathcal{N}[\phi^{(n)}_i]| \leq \mathcal{O}(1)$ for $n \geq 3$ hold with a generic odd function $f(\bullet)$ (see Remark in Section 4.2 for details), $\Cov_\mathcal{N}[\phi_i, \phi_j]$ is order $N_s^{-3/2}$. Whereas, the $i$-th diagonal element $\Cov_\mathcal{N}[\phi_i,\phi_i]$ is order 1. These observations conclude that eigenvalues of $\Cov_\mathcal{N}[\phi]$---and therefore those of $\Sigma$ according to Lemma 1 (see Remark in Section 4.1 for details)---are not larger than order $\max[1,N_fN_s^{-3/2}]$; thus, all nonzero eigenvalues of $HH^T$---which are order $N_f/N_s$---are sufficiently greater than those of $\Sigma$ when $N_f \gg N_s \gg 1$.

One can further proceed the calculation of $\Sigma$ by explicitly computing the coefficients $\E_\mathcal{N}[\phi^{(n)}_i]$ up to the fourth order. Because $f(\bullet)$ is an odd function, using $\Cov_\mathcal{N}[y_i, y_j]^n = ((AA^T)^{\odot n})_{ij}$, equation \eqref{Cov_phi} becomes $\Cov_\mathcal{N}[\phi_i, \phi_j] = \E_\mathcal{N}[f^{(3)}(y_i)]\E_\mathcal{N}[f^{(3)}(y_j)] \{ a_i a_j((AA^T)^{\odot 2})_{ij} /2 + ((AA^T)^{\odot 3})_{ij} /6 \} + \mathcal{O}(N_s^{-5/2})$ (see Remark in Section 4.2 for details). This analytical expression involves the Hadamard (element-wise) power of matrix $AA^T$ (denoted by $\odot$), e.g., $(AA^T)^{\odot 3} \equiv (AA^T) \odot (AA^T) \odot (AA^T)$. When $N_f > N_s^2 \gg 1$, $(AA^T)^{\odot 3}$ has $N_s$ major eigenvalues---all of which are $3N_f/N_s^2$ (as the leading order) with the corresponding eigenvectors that match the directions of $AA^T$---and $(N_f-N_s)$ minor eigenvalues that are negligibly smaller than order $N_f/N_s^2$ (Lemma 3 in Methods). This property yields an approximation $(AA^T)^{\odot 3} - \diag[(AA^T)^{\odot 3}] = (3/N_s)(AA^T - \diag[AA^T])$ up to the negligible minor eigenmodes. Note that when $N_s^2 > N_f \gg 1$, off-diagonal elements of $(AA^T)^{\odot 3}$ are negligible compared to diagonal elements of $\Sigma$ (see below). Similarly, $(AA^T)^{\odot 2}$ has one major eigenvalue $N_f/N_s$ in the direction of $(1,...,1)^T$, while other minor eigenmodes of it are negligible. These approximations are used to compute off-diagonal elements of $\Sigma$.

Hence, $\Sigma$ is analytically expressed as
\begin{equation}
\label{Sigma}
\Sigma = \left( \overline{f^2} - \overline{f'}^2 \right)I + \frac{\overline{f^{(3)}}^2}{2N_s}(AA^T + aa^T) + \Xi.
\end{equation}
Here, from $\Sigma = \Cov[f] - HH^T$ and equation \eqref{H}, each diagonal element is expressed as $\Sigma_{ii} = \overline{f^2} - \overline{f'}^2 + \mathcal{O}(N_s^{-1/2})$ using $\overline{f^2} \equiv \int f^2(\xi)\exp(-\xi^2/2){\rm d}\xi/\sqrt{2\pi}$, which yields the first term of equation \eqref{Sigma} (where the error term is involved in the third term). The second term is generated from equation \eqref{Cov_phi} followed by Lemma 3, where $\E_\mathcal{N}[f^{(3)}(y_i)]$ is approximated by $\overline{f^{(3)}} \equiv \int f^{(3)}(\xi)\exp(-\xi^2/2){\rm d}\xi/\sqrt{2\pi}$ up to order $N_s^{-1/2}$. The third term is the error matrix $\Xi$ that summarizes the deviations caused by the non-Gaussianity of $p(y_i,y_j)$ (c.f., Lemma 1; see also Remark in Section 4.1), higher order terms of the Taylor series in equation \eqref{Cov_phi}, minor eigenmodes of $(AA^T)^{\odot 2}$ and $(AA^T)^{\odot 3}$, and the effect of the order $N_s^{-1/2}$ deviations of coefficients (e.g., $\E[f_i^2]$) from their approximations (e.g., $\overline{f^2}$). Due to its construction, eigenvalues of $\Xi$ are smaller than those of the first or second term of equation \eqref{Sigma}. This indicates that for large $N_s$, either the first or second term of equation \eqref{Sigma} provides the largest eigenvalue of $\Sigma$, which is order $\max[1,N_f/N_s^2]$. Thus, $\Xi$ is negligible for the following analyses.


Accordingly, all nonzero eigenvalues of $HH^T$---that are order $N_f/N_s$---are much greater than the maximum eigenvalue of $\Sigma$---that is order $\max[1,N_f/N_s^2]$---when $N_f \gg N_s \gg 1$. Thus, unless $B$ specifically attenuates one of $N_s$ major eigenmodes of $HH^T$ or significantly amplifies a particular eigenmode of $\Sigma$, we have
\begin{equation}
\label{eig_ratio}
\min\eig[H^TB^TBH] \gg \max\eig[B\Sigma B^T]
\end{equation}
for $N_x = N_f \gg N_s \gg 1$. Here, $\eig[\bullet]$ indicates a set of eigenvalues of $\bullet$. Because $\min\eig[H^TB^TBH]$ is not smaller than order $N_f/N_s \min\eig[B^TB]$, while $\max\eig[B\Sigma B^T]$ is not larger than order $\max[1,N_f/N_s^2] \max\eig[B^TB]$, inequality \eqref{eig_ratio} holds at least when $N_f/N_s$ and $N_s$ are much greater than $\max\eig[B^TB]/\min\eig[B^TB]$. This is the case, for example, if all the singular values of $B$ are order 1. We call $B$ is sufficiently isotropic when $\max\eig[B^TB]/\min\eig[B^TB]$ can be upper-bounded by a (possibly large) finite constant, and focus on such $B$; as otherwise, the effective input dimensionality is much smaller than $N_x$. In other words, one can redefine $N_x$ of any system by replacing the original sensory inputs with their compressed representation to render $B$ isotropic.

When inequality \eqref{eig_ratio} holds, the first $N_s$ major eigenmodes of $\Cov[x]$ precisely match the signal covariance, while minor eigenmodes are negligibly small. Hence, there is a clear spectrum gap between the largest $N_s$ eigenvalues and the rest. This indicates that one can reliably identify the signal covariance and source dimensionality based exclusively on PCA of $\Cov[x]$ in an unsupervised manner. Hence, when $\Lambda_M \in \mathbb{R}^{N_s \times N_s}$ is a diagonal matrix that arranges the first $N_s$ major eigenvalues of $\Cov[x]$ in descending order, with the corresponding eigenvector matrix $P_M \in \mathbb{R}^{N_x \times N_s}$, we obtain
\begin{equation}
\label{major_components}
P_M\Lambda_M P_M^T \simeq BHH^TB^T.
\end{equation}
The corrections of $\Lambda_M$ and $P_M$ due to the presence of the residual covariance is estimated as follows: by the first order perturbation theorem (Griffiths, 2005), the correction of the $i$-th major eigenvalue of $\Cov[x]$ is upper-bounded by $\max\eig[B\Sigma B^T]$, while the norm of the correction of $i$-th eigenvector is upper-bounded by $\max\eig[B\Sigma B^T] / \min\eig[H^TB^TBH]$. These corrections are negligibly smaller than the leading order terms, when inequality \eqref{eig_ratio} holds. Further details are provided in Methods.

Therefore, applying PCA to $\Cov[x]$ can extract the signal covariance as the major principal components with high accuracy when $N_x = N_f \gg N_s \gg 1$ and $B$ is sufficiently isotropic. This property is numerically validated. The major principal components of $\Cov[x]$ suitably approximate the signal covariance (Fig. 2A). In the simulations, elements of $B$ are sampled from a Gaussian distribution with zero mean. The trace ratio $\tr[B\Sigma B^T]/\tr[H^TB^TBH]$ (i.e., the ratio of the eigenvalue sum of residual covariance $\tr[B\Sigma B^T]$ to that of signal covariance $\tr[H^TB^TBH]$) retains the value of about 0.6 irrespective of $N_s$ and $N_x$, indicating that the large scale systems we consider are fairly nonlinear (Fig. 2B). In contrast, the eigenvalue ratio $\max\eig[B\Sigma B^T] / \min\eig[H^TB^TBH]$ monotonically converges to zero when $N_x$ increases; thus, inequality \eqref{eig_ratio} holds for large $N_x$ (Fig. 2C). Consequently, $P_M$ approximates nonzero eigenmodes of the signal covariance accurately (Fig. 2D). These results indicate that PCA is a promising method for reliably identifying the linear components in sensory inputs. In what follows, we explicitly demonstrate that projecting the inputs onto the major eigenspace can recover true hidden sources of sensory inputs accurately.


\begin{figure}[H]
  \centering
  \includegraphics[width=1.00\linewidth]{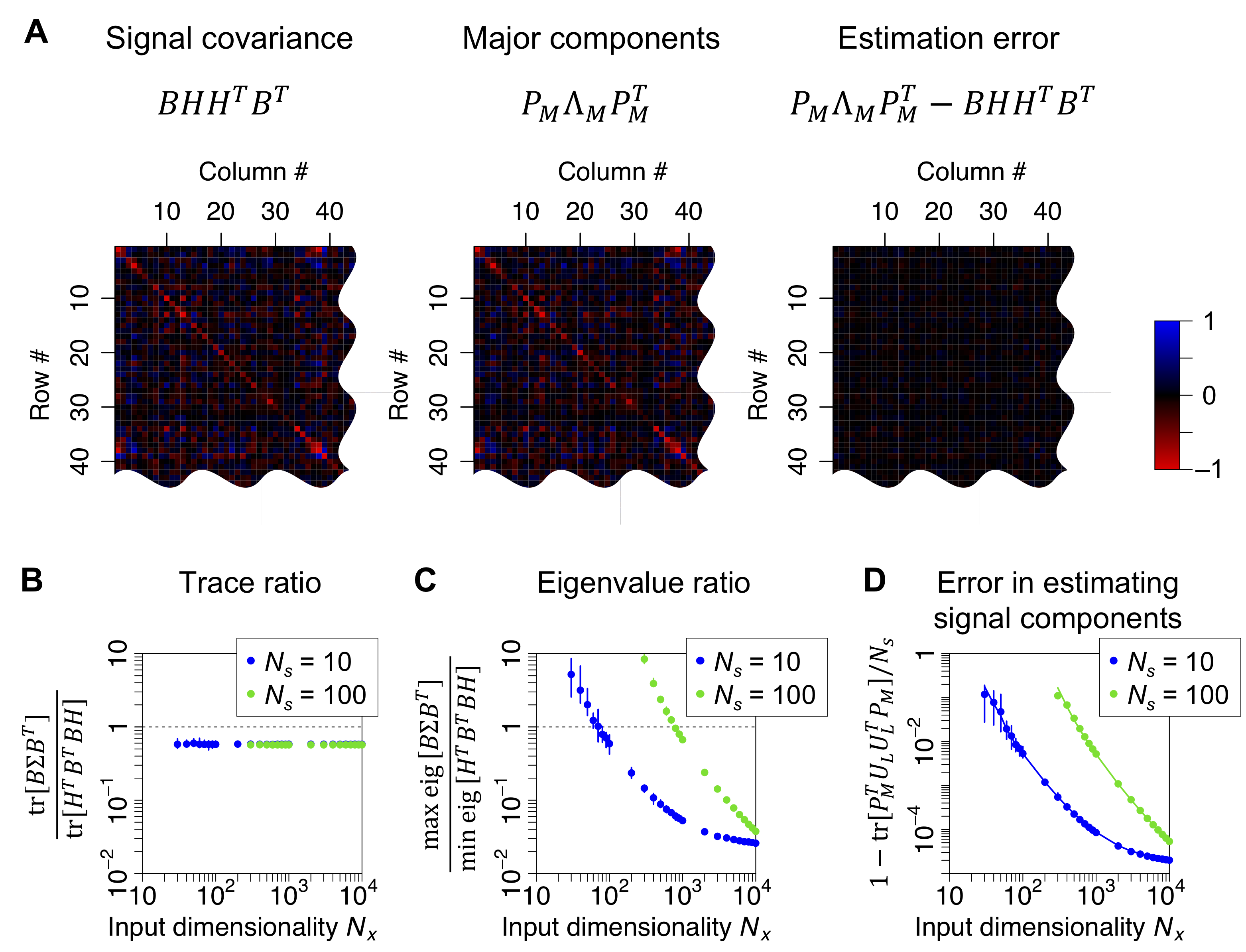}
\end{figure}
\noindent Figure 2. PCA can identify linear components comprising linear projections from every true hidden source. Here, hidden sources $s$ are independently generated from an identical uniform distribution with zero mean and unit variance; $N_f=N_x$ and $f(\bullet) = \sign(\bullet)$ are fixed; elements of $A,a$ are independently sampled from $\mathcal{N}[0,1/N_s]$; and elements of $B$ are independently sampled from $\mathcal{N}[0,1/N_f]$. PCA is achieved via eigenvalue decomposition. (\textbf{A}) Comparison between the signal covariance and major principal components of sensory inputs, when $N_s=10$ and $N_x=10^3$. (\textbf{B}) Trace ratio $\tr[B\Sigma B^T]/\tr[H^TB^TBH]$. (\textbf{C}) Eigenvalue ratio $\max\eig[B\Sigma B^T] / \min\eig[H^TB^TBH]$. (\textbf{D}) Error in extracting nonzero eigenmodes of signal covariance as major components, which is scored by $1-\tr[P_M^TU_LU_L^TP_M]/N_s$, where $U_L \in \mathbb{R}^{N_x \times N_s}$ indicates the left-singular vectors of $BH$. As $N_x$ increases, $P_M P_M^T$ converges to $U_L U_L^T$ reliably and accurately. Solid lines indicate theoretical values of estimation errors: see equation \eqref{eigenvalue_estimator} in Methods for details. Circles and error bars indicate the means and areas between maximum and minimum values obtained with 20 different realizations of $s,A,B,a$, where some error bars are hidden by the circles. \\


\subsection{Asymptotic linearization theorem}
We now consider a linear encoder comprising a single-layer linear neural network,
\begin{equation}
\label{u}
u \equiv W \left(x-\E[x]\right),
\end{equation}
where $u \equiv (u_1, \dots, u_{N_s})^T$ are $N_s$-dimensional neural outputs, and $W \in \mathbb{R}^{N_s \times N_x}$ is a synaptic weight matrix. Suppose that by applying PCA to the inputs, one obtains $W$ that represents a subspace spanned by the major eigenvectors, $W = \Omega_M \Lambda_M^{-1/2} P_M^T$, where $\Omega_M$ is an arbitrary $N_s \times N_s$ orthogonal matrix expressing an ambiguity. A normalization factor $\Lambda_M^{-1/2}$ is multiplied such as to ensure $\Cov[u] = I$. Neural outputs with this $W$ indeed express the {\it optimal} linear encoder of the inputs because it is the solution of the maximum likelihood estimation that minimizes the loss to reconstruct the inputs from lower-dimensional encoder $u$ using a linear network under Gaussian assumption; see e.g., (Xu, 1993; Wentzell et al., 1997) for related studies. In other words, when we assume that the loss follows a unit Gaussian distribution, $\arg\min_W \E[|x - \E[x] - W^+ u|^2] = \Omega_M \Lambda_M^{-1/2} P_M^T$ holds under the constraint of $\dim[u] = N_s$ and $\Cov[u] = I$, where $W^+$ indicates the pseudo inverse of $W$.

Crucially, equation \eqref{major_components} directly provides the key analytical expression to represent a subspace spanned by the linear components:
\begin{equation}
\label{W}
W \simeq \Omega (BH)^+.
\end{equation}
Here, $(BH)^+ \equiv (H^TB^TBH)^{-1}H^TB^T$ is the pseudo inverse of $BH$ and $\Omega$ is another arbitrary $N_s \times N_s$ orthogonal matrix. Error in approximating $(BH)^+$ is negligible for the following calculations as long as inequality \eqref{eig_ratio} holds (see Methods for more details). Equation \eqref{W} indicates that the directions of the linear components can be computed under the BSS setup---up to an arbitrary orthogonal ambiguity factor $\Omega$. Hence, we obtain the following theorem:


\paragraph{\textit{\textbf{Theorem (asymptotic linearization)}}}
{\it When inequality \eqref{eig_ratio} holds, from equations \eqref{x2}, \eqref{u}, and \eqref{W}, the linear encoder with optimal matrix $W = \Omega_M \Lambda_M^{-1/2} P_M^T$ can be analytically expressed as}
\begin{equation}
u = \Omega (s + \varepsilon)
\end{equation}
{\it using an linearization error $\varepsilon \equiv (BH)^+ B\phi$ with the covariance matrix of}
\begin{equation}
\label{Cov_eps}
\Cov[\varepsilon] = (BH)^+ B\Sigma B^T(BH)^{+T}.
\end{equation}
{\it The maximum eigenvalue of $\Cov[\varepsilon]$ is upper-bounded by $\max\eig[B\Sigma B^T]/\min\eig[H^TB^TBH]$, which is sufficiently smaller than 1 from inequality \eqref{eig_ratio}. In particular, when $p_s(s)$ is a symmetric distribution, $f(\bullet)$ is an odd function, and $B$ is sufficiently isotropic, using equation \eqref{Sigma}, $\Cov[\varepsilon]$ can be explicitly computed as}
\begin{equation}
\label{Cov_eps2}
\Cov[\varepsilon] = \frac{N_s}{N_f}\left( \frac{\overline{f^2}}{\overline{f'}^2} - 1 \right) (I + \Delta) + \frac{\overline{f^{(3)}}^2}{2N_s \overline{f'}^2} I
\end{equation}
{\it as the leading order. Symmetric matrix $\Delta \equiv U_A^T (B^TB-I)^2 U_A$ characterizes the anisotropy of $B^TB$ in the directions of the left-singular vectors of $A$, $U_A \in \mathbb{R}^{N_f \times N_s}$, wherein $\max\eig[\Delta]$ is upper-bounded by $\max\eig[(B^TB-I)^2] = \mathcal{O}(1)$. Together, we conclude that}
\begin{equation}
u = \Omega s + \mathcal{O}\left( \sqrt{ \frac{N_s}{N_f} } \right) + \mathcal{O}\left( \frac{1}{\sqrt{N_s}} \right).
\end{equation}
\hspace{10pt}


The derivation detail of equation \eqref{Cov_eps2} is provided in Methods. Equation \eqref{Cov_eps2} indicates that the linearization error monotonically decreases as $N_f/N_s$ and $N_s$ increase; thus, $u$ converges to a linear mixture of all the true hidden sources ($u \to \Omega s$) in the limit of large $N_f/N_s$ and $N_s$. Only the anisotropy of $B^TB$ in the directions of $U_A$ increases the linearization error. In essence, applying PCA to the inputs effectively filters out nonlinear components in the inputs because the majority of nonlinear components is perpendicular to the directions of the signal covariance.

Although the obtained encoder is not independent of each other because of the multiplication with $\Omega$, it is remarkable that the proposed approach enables the conversion of the original nonlinear BSS problem to a simple linear BSS problem. This indicates that $u$ can be separated into each independent encoder by further applying a linear ICA method (Comon, 1994; Bell \& Sejnowski, 1995; Bell \& Sejnowski, 1997; Amari et al., 1996; Hyvarinen \& Oja, 1997) to it, and these independent encoders match the true hidden sources up to permutations and sign-flips. Zero-element-wise-error nonlinear BSS is attained in the limit of large $N_f/N_s$ and $N_s$. As a side note, PCA can indeed recover a linear transformation of true hidden sources in the inputs even when these sources have higher-order correlations if the average of these correlations converges to zero with high source dimensionality. This property is potentially useful, for example, for identifying true hidden states of time series data generated from nonlinear systems (Isomura \& Toyoizumi, 2020).

In summary, we analytically quantified the accuracy of the optimal linear encoder---obtained through the cascade of PCA and ICA---in inverting the nonlinear generative process to identify all hidden sources. The encoder increases its accuracy as $N_f/N_s$ and $N_s$ increase, and asymptotically attains the true hidden sources.

The proposed theorem is empirically validated by numerical simulations. Each element of the optimal linear encoder obtained by the PCA--ICA cascade represents a hidden source, wherein BSS errors (difference between true and estimated sources) decrease as $N_x$ increases (Fig. 3A). The PCA--ICA cascade performs the nonlinear BSS with various types of nonlinear basis functions (Figs. 3B and 3C). This is a remarkable property of the PCA--ICA cascade because these results indicate that it can perform the nonlinear BSS without knowing the true nonlinearity that characterizes the generative process. Although equation \eqref{Cov_eps2} is not applicable to non-odd nonlinear basis function $f(\bullet)$, empirical observations indicate that the PCA--ICA cascade can identify the true hidden sources even with non-odd function $f(\bullet)$ (Fig. 3C), as long as $H = \E[fs^T]$ is nonzero.

The log-log plot illustrates that the magnitude of element-wise BSS errors---scored by the mean squared error---decreases inversely proportional to $N_x/N_s$; however, it saturates around $N_x = N_s^2$ (Fig. 3D). These observations validate equation \eqref{Cov_eps2} which asserts that the linearization error is determined by the sum of $N_s/N_f$ and $1/N_s$ order terms. Although equation \eqref{Cov_eps2} overestimates the BSS error when $N_s = 10$, this is because each dimension of $As$ significantly deviates from a Gaussian variable due to small $N_s$---as $s$ is sampled from a uniform distribution in these simulations. We confirm that when hidden sources are generated from a distribution close to Gaussian, actual BSS errors shift toward the theoretical value of equation \eqref{Cov_eps2} even when $N_s = 10$ (cross marks in Fig. 3D). Indeed, this deviation disappears for large $N_s$ according to the central limit theorem. Therefore, equation \eqref{Cov_eps2} is a good approximation of actual BSS errors for $N_x = N_f \gg N_s \gg 1$, and the proposed theorem suitably predicts the performance of the PCA--ICA cascade for a wide range of nonlinear BSS setups.


\begin{figure}[H]
  \centering
  \includegraphics[width=1.00\linewidth]{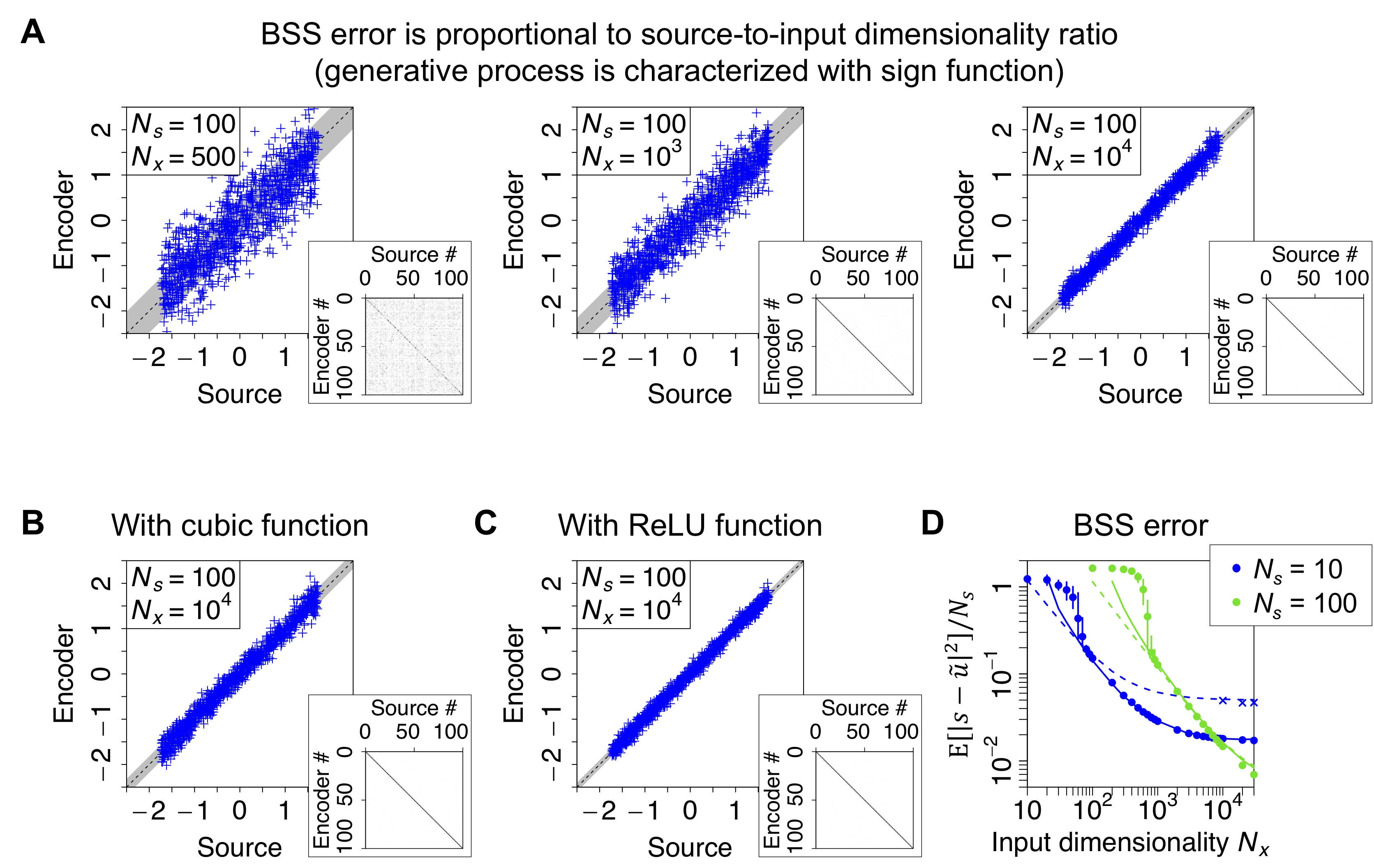}
\end{figure}
\noindent Figure 3. BSS error can be characterized by the source and input dimensionalities. In all panels, $s$ is generated by an identical uniform distribution; $N_f=N_x$ is fixed; and $A,B,a$ are sampled from Gaussian distributions as in Fig. 2. Encoders $\tilde{u} = (\tilde{u}_1, ..., \tilde{u}_{N_s})^T$ are obtained by applying the PCA--ICA cascade to sensory inputs. For visualization purpose, elements of $\tilde{u}$ are permuted and sign-flipped to ensure that $\tilde{u}_i$ encodes $s_i$. \textbf{(A)} Each element of $\tilde{u}$ encodes a hidden source, wherein an error in representing a source monotonically decreases when $N_x$ increases. The generative process is characterized with $N_s=100$ and $f(\bullet) = \sign(\bullet)$. Shaded areas represent theoretical values of the standard deviation computed using equation \eqref{Cov_eps2}. As elements of $B$ are Gaussian distributed, $\Delta = I$ holds. Inset panels depict the absolute value of covariance matrix $|\Cov[\tilde{u},s]|$ with grayscale values ranging from 0 (white) to 1 (black). A diagonal covariance matrix indicates the successful identification of all the true hidden sources. \textbf{(B)(C)} Nonlinear BSS when the generative process is characterized by $f(\bullet) = (\bullet)^3$ or $f(\bullet) = \mathrm{ReLU}(\bullet)$. $\mathrm{ReLU}(\bullet)$ outputs $\bullet$ for $\bullet>0$ or 0 otherwise. In (C), the shaded area is computed using equation \eqref{Cov_eps}. \textbf{(D)} Quantitative relationship between the source and input dimensionalities and element-wise BSS error scored by the mean squared error $\E[|s-\tilde{u}|^2]/N_s$. Here, $f(\bullet) = \sign(\bullet)$ is supposed. Circles and error bars indicate the means and areas between maximum and minimum values obtained with 20 different realizations of $s,A,B,a$, where some error bars are hidden by the circles. Solid and dashed lines represent theoretical values computed using equations \eqref{Cov_eps} and \eqref{Cov_eps2}, respectively. These lines fit the actual BSS errors when $N_x \gg N_s \gg 1$, although some deviations occur when $N_s$ or $N_x$ is small. Blue cross marks indicate actual BSS errors for $N_s=10$ when hidden sources are sampled from a symmetric truncated normal distribution. \\


\subsection{Hebbian-like learning rules can reliably solve nonlinear BSS problem}
As a corollary of the proposed theorem, a linear neural network can identify true hidden sources through Hebbian-like plasticity rules in the nonlinear BSS setup under consideration. Oja's subspace rule (Oja, 1989)---which is a modified version of the Hebbian plasticity rule (see Methods)---is a well-known PCA approach that extracts the major eigenspace without yielding a spurious solution or attaining a local minimum (Baldi \& Hornik, 1989; Chen et al., 1998). Thus, with generic random initial synaptic weights, this Hebbian-like learning rule can quickly and reliably identify an optimal linear encoder that can recover true hidden sources from their nonlinearly mixed inputs in a self-organizing or unsupervised manner.

Numerical experiments demonstrate that regardless of the random initialization of synaptic weight matrix $W_{\rm PCA} \in \mathbb{R}^{N_s \times N_x}$, Oja's subspace rule updates $W_{\rm PCA}$ to converge to the major eigenvectors, i.e., the directions of the linear components. The accuracy of extracting the linear components increases as the number of training samples increases, and reaches the same accuracy as an extraction via eigenvalue decomposition (Fig. 4A). Because the original nonlinear BSS problem has been now transformed to a simple linear BSS problem, the following linear ICA approach (Comon, 1994; Bell \& Sejnowski, 1995; Bell \& Sejnowski, 1997; Amari et al., 1996; Hyvarinen \& Oja, 1997) can reliably separate all the hidden sources from the features extracted using Oja’s subspace rule. Amari's ICA algorithm (Amari et al., 1996)---which is another Hebbian-like rule (see Methods)---updates synaptic weight matrix $W_{\rm ICA} \in \mathbb{R}^{N_s \times N_s}$ to render neural outputs independent of each other. The obtained independent components accurately match the true hidden sources of the nonlinear generative process up to their permutations and sign-flips (Fig. 4B). These results highlight that the cascade of PCA and ICA---implemented via Hebbian-like learning rules---can self-organize the optimal linear encoder and therefore identify all the true hidden sources in this nonlinear BSS setup, with high accuracy and reliability when $N_x = N_f \gg N_s \gg 1$.


\begin{figure}[H]
  \centering
  \includegraphics[width=0.70\linewidth]{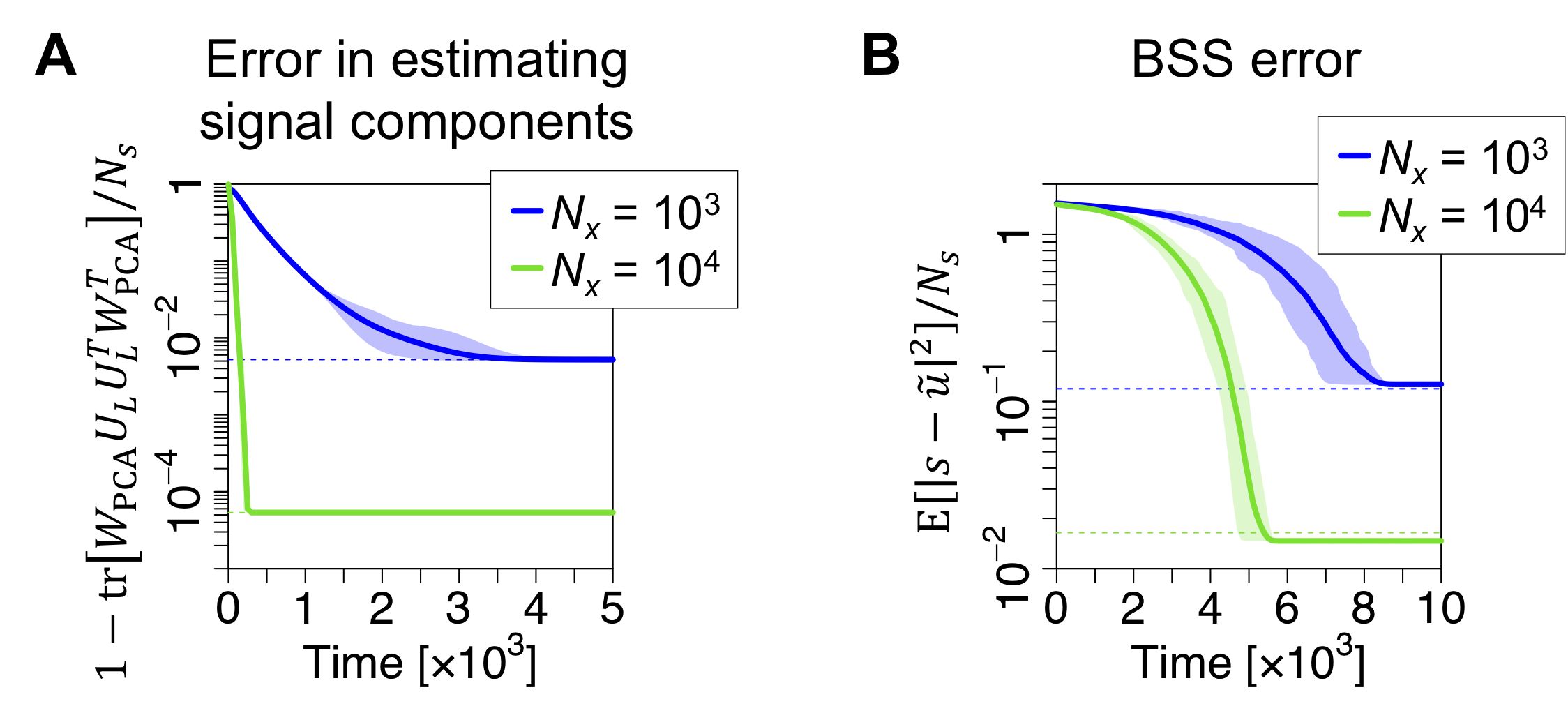}
\end{figure}
\noindent Figure 4. Hebbian-like learning rules can identify the optimal linear encoder. As in Figs. 2 and 3, $s$ is generated by an identical uniform distribution; $N_s = 100$, $N_f = N_x$, and $f(\bullet) = \sign(\bullet)$ are fixed; and $A,B,a$ are sampled from Gaussian distributions. \textbf{(A)} Learning process of Oja's subspace rule for PCA. Synaptic weight matrix $W_{\rm PCA} \in \mathbb{R}^{N_s \times N_x}$---initialized as a random matrix---converges to $P_M^T$ up to multiplication of an orthogonal matrix from the left-hand side. This indicates that Oja's rule extracts the linear components according to the proposed theorem. Dashed lines indicate the estimation errors when eigenvalue decomposition is applied (c.f., Fig. 2C). \textbf{(B)} Learning process of Amari's ICA algorithm. Synaptic weight matrix $W_{\rm ICA} \in \mathbb{R}^{N_s \times N_s}$---initialized as a random matrix---learns to separate the compressed inputs $W_{\rm PCA}(x-\E[x])$ into independent signals. Because PCA yields a linear transformation of hidden sources, the ensuing independent encoder $\tilde{u} \equiv W_{\rm ICA} W_{\rm PCA} (x - \E[x])$ identifies all the true hidden sources with a small BSS error. Elements of $\tilde{u}$ are permuted and sign-flipped to ensure that $\tilde{u}_i$ encodes $s_i$. Dashed lines are computed using equation \eqref{Cov_eps2}. In (A) and (B), learning rates are $\eta_{\rm PCA} = 10^{-3}$ and $\eta_{\rm ICA} = 0.02$, respectively. Solid lines represent the mean estimation errors, while shaded areas represent areas between maximum and minimum values obtained with 20 different realizations of $s,A,B,a$. \\


\section{Discussion}
In this work, we theoretically quantified the accuracy of nonlinear BSS performed using the cascade of linear PCA and ICA when sensory inputs are generated from a two-layer nonlinear generative process. First, we demonstrated that as the dimensionality of hidden sources increases and the dimensionalities of sensory inputs and nonlinear bases increase relative to the source dimensionality, the first $N_s$ major principal components approximately express a subspace spanned by the linear projections from all hidden sources of the sensory inputs. Under the same condition, we then demonstrated that the optimal linear encoder obtained by projecting the inputs onto the major eigenspace can accurately recover all the true hidden sources from their nonlinear mixtures. This property is termed the asymptotic linearization theorem. The accuracy of the subspace extraction increases as $N_f/N_s$ and $N_s$ increase, because the gap between the minimum eigenvalue of the linear (i.e., signal) components and maximum eigenvalue of the nonlinear (i.e., residual) components becomes significantly large. Hebbian-like plasticity rules can also identify the optimal linear encoder by extracting major principal components in a manner equal to PCA. Subsequent application of linear ICA on the extracted principal components can reliably identify all the true hidden sources up to permutations and sign-flips. Unlike conventional nonlinear BSS methods that can yield spurious solutions (Hyvarinen \& Pajunen, 1999; Jutten \& Karhunen, 2004), the PCA--ICA cascade is guaranteed to identify the true hidden sources in the asymptotic condition because it successfully satisfies Requirements 1--5 specified earlier.

The unique identification of true hidden sources $s$ (up to permutations and sign-flips) is widely recognized only under the linear BSS setup. In contrast, it is well-known that conventional nonlinear BSS approaches using nonlinear neural networks do not guarantee the identification of true hidden sources under the general nonlinear BSS setup (Hyvarinen \& Pajunen, 1999; Jutten \& Karhunen, 2004). One may ask if nonlinear BSS methods may find a component-wise nonlinear transformation of the original sources $s' = g(s)$ instead of $s$ because $s'$ is still an independent source representation. This is observed when conventional nonlinear BSS approaches based on the nonlinear ICA are employed because the nonlinear ICA finds one of many representations that minimize the dependency among outputs. However, such nonlinear BSS approaches do not guarantee the identification of true sources. This is because when $g'(s)$ is a component-wise nonlinear transformation that renders a non-Gaussian source Gaussian distributed, a transformation $s'' = g(Rg'(s))$---characterized by some orthogonal matrix $R$ and component-wise nonlinear transformation $g$---can yield an arbitrary independent representation $s''$ that differs from the original sources. The former ($s'$) is a special case of the latter ($s''$). Even in the former case, finding $s'$ transformed by a highly nonlinear, non-monotonic function $g(s)$ is problematic. Thus, the general nonlinear BSS is essentially an ill-posed problem. Having said this, earlier works typically investigated the case wherein the sources and inputs have the same dimensionality, while the other conditions have not been extensively analyzed. Thus, in the current work, we focused on the nonlinear BSS under the condition where $N_x = N_f \gg N_s \gg 1$.

A remarkable aspect of the proposed nonlinear BSS approach is that it utilizes the linear PCA to extract the linear components of the original sources from the mixed sensory inputs. This is the process that enables the reduction of the original nonlinear BSS to a simple linear BSS and consequently ensures the reliable identification of the true sources. In other words, the proposed approach does not rely on the nonlinear ICA; thus, no concerns exist regarding the creation of the aforementioned spurious solutions. We mathematically demonstrated the absence of such spurious solutions when $N_x = N_f \gg N_s \gg 1$. Specifically, we adopted a minimal assumption about the relationship between $B$ and the system dimensionalities such that $N_f/N_s$ and $N_s$ are much greater than $\max\eig[B^TB]/\min\eig[B^TB]$. Our mathematical analyses demonstrate that this condition is sufficient to asymptotically determine the possible form of hidden sources---that can generate the observed sensory inputs $x$---in a unique manner. In contrast, in the limit of large $N_f/N_s$ and $N_s$, no nonlinear transformation of $s$ (i.e., $s'$ or $s''$) can generate the observed sensory inputs while satisfying the aforementioned condition \footnote{This property can be understood as follows: because $x$ is generated through $x=Bf(As+a)$ with a sufficiently isotropic $B$ that satisfies inequality \eqref{eig_ratio}, from the proposed theorem, the encoder $u=\Lambda_M^{-1/2} P_M^T (x-\E[x])=\Omega(s+\varepsilon)$ asymptotically becomes $u\to\Omega s$ in the limit of large $N_f/N_s$ and $N_s$. In addition, we define $s'=g(s)\in\mathbb{R}^{N_s}$ as a nonlinear transformation of $s$. If there is another generative process $x=B'f'(A's'+a')$ that can generate the observed $x$ while satisfying inequality \eqref{eig_ratio}---where $A'$ and $a'$ are Gaussian distributed matrix and vector, $f'$ is a nonlinear function (not the derivative of $f$, unlike in the main text), and $B'$ is a sufficiently isotropic matrix---the encoder can also be expressed as $u'=\Lambda_M^{-1/2} P_M^T (x-\E[x])=\Omega'(s'+\varepsilon')$, which asymptotically becomes $u'\to\Omega's'$ in the limit of large $N_f/N_s$ and $N_s$. Note that $\Lambda_M$ and $P_M$ are the same as above because $x$ does not change. However, while $u\equiv u'$ holds by construction, $\Omega s\not\equiv\Omega's'$ for any $\Omega$ and $\Omega'$ because $s'$ is a nonlinear transformation of $s$. Thus, such a generative process $x=B'f'(A's'+a')$ does not exist in the limit of large $N_f/N_s$ and $N_s$.}. Thus, when inequality \eqref{eig_ratio} holds, the true hidden sources $s$ are uniquely determined up to permutations and sign-flips ambiguities without component-wise nonlinear ambiguity. (Conversely, if inequality \eqref{eig_ratio} does not hold, it might not be possible to distinguish true hidden sources and their nonlinear transformations in an unsupervised manner.) Hence, under the condition we consider, the nonlinear BSS is formally reduced to a simple linear BSS, wherein only permutations and sign-flips ambiguities exist while no nonlinear ambiguity remains. This property is crucial for identifying the true sources under the nonlinear BSS setup, without being attracted by spurious solutions such as $s'$ and $s''$.

The nonlinear generative processes that we considered in this work are sufficiently generic. Owing to the universality of two-layer networks, each element of an arbitrary generative process $x=F(s)$ can be approximated using equation \eqref{x} with a high degree of accuracy in the component-wise sense when $N_x = N_f \gg N_s \gg 1$ \footnote{It should be noted that, unlike the condition we considered (i.e., $N_x = N_f \gg N_s \gg 1$), when $N_f \gg N_x$, one may observe a counterexample of the proposed theorem such that the major principal components are dissimilar to a linear transformation of the true hidden sources, as is well-known in the literature on nonlinear ICA (Hyvarinen \& Pajunen, 1999; Jutten \& Karhunen, 2004). This is because when $N_f \gg N_x$, a general $B$ may not be sufficiently isotropic. In other words, $N_f \gg N_x$ corresponds to the case wherein $N_x$ is insufficient to ensure inequality \eqref{eig_ratio}; thus, $N_x$ needs to be greater. Hence, it is remarkable that the proposed theorem specifies the condition under which the achievability of the nonlinear BSS is mathematically guaranteed.}. This allows one to analyze the achievability of the nonlinear BSS using a class of generative processes comprising random basis functions, as a proxy for investigating $x=F(s)$. As a general property of such generative processes, the eigenvalues of the signal covariance (i.e., linear components) become significantly larger than those of the residual covariance (i.e., nonlinear components) when $N_x = N_f \gg N_s \gg 1$. Sufficient system dimensionalities to exhibit this property are determined depending on matrix $B$ that characterizes the generative process. Namely, this observation holds true at least when both $N_f/N_s$ and $N_s$ are much greater than $\max\eig[B^TB]/\min\eig[B^TB]$. The existence of such $N_s$ and $N_f$ is guaranteed if we consider sufficiently isotropic $B$ wherein $\max\eig[B^TB]/\min\eig[B^TB]$ is upper-bounded by a large finite constant. Meanwhile, with such a large constant, the two-layer networks can approximate each element of $x=F(s)$ accurately. This in turn means that, as $N_f/N_s$ and $N_s$ increase, the proposed theorem becomes applicable to a sufficiently broad class of nonlinear generative processes.

This asymptotic property can be understood as follows: when adding a new random basis ($f_k=H_k s+\phi_k$) to the existing large-scale system, the linear component of $f_k$ ($H_k s$) must be placed within the existing low-dimensional subspace spanned only by $N_s$ ($\ll N_f$) linear projections of sources. In contrast, its nonlinear component ($\phi_k$) is almost uncorrelated with other nonlinear components ($\phi$), as shown in equation (2.6), because these components are characterized by Gaussian distributed $A$. Thus, the former increases the eigenvalues of the signal covariance in proportion to $N_f/N_s$, while the latter adds a new dimension in the subspace of nonlinear components without significantly increasing the maximum eigenvalue of the residual covariance. Owing to this mechanism, the eigenvalues of the signal covariance become $N_f/N_s$ times larger than those of the residual covariance when $N_s \gg 1$. Hence, the linear PCA is sufficient to reduce the nonlinear BSS to a linear BSS.

Nonlinear variants of PCA, such as autoencoders (Hinton \& Salakhutdinov, 2006; Kingma \& Welling, 2013), have been widely used for representation learning. Because natural sensory data are highly redundant and do not uniformly cover the entire input space (Chandler \& Field, 2007), finding a concise representation of the sensory data is essential to characterize its properties (Arora, 2017). In general, if a large nonlinear neural network is used, many equally good solutions will be produced (Kawaguchi, 2016; Lu \& Kawaguchi, 2017; Nguyen \& Hein, 2017). In this case, there is no objective reason to select one solution over another if they have similar reconstruction accuracy. However, this property also leads to infinitely many spurious solutions if the aim is the identification of the true hidden sources (Hyvarinen \& Pajunen, 1999; Jutten \& Karhunen, 2004). Consequently, the outcomes of these approaches using nonlinear neural networks are intrinsically ambiguous, and obtained solutions highly depend on the heuristic design of the regularization parameters used in the networks and learning algorithms (Dahl et al., 2012; Hinton et al., 2012; Wan et al., 2013). However, unlike those nonlinear approaches, we prove that the cascade of linear PCA and ICA ensures that the true hidden sources of sensory inputs are obtained if there are sufficient dimensionalities in mappings from hidden sources to sensory inputs. Thus, this linear approach is suitable to solve the nonlinear BSS problem while retaining the guarantee to identify the true solution. Its BSS error converges in proportion with the source-to-input dimensionality ratio for a large-scale system comprising high-dimensional sources. Furthermore, if needed, its outcomes can be used as a plausible initial condition for nonlinear methods to further learn the forward model of the generative process.

Because most natural data (including biological, chemical, and social data) are generated from nonlinear generative processes, the scope of application of nonlinear BSS is quite broad. The proposed theorem provides a theoretical justification for the application of standard linear PCA and ICA to nonlinearly generated natural data to infer their true hidden sources. An essential lesson from the proposed theorem is that an artificial intelligence should employ a sufficient number of sensors to identify true hidden sources of sensory inputs accurately, because the BSS error is proportional to the source-to-input dimensionality ratio in a large-scale system.

In the case of computer vision, low-dimensional representations (typically 10--$10^4$ components) are extracted from up to millions of pixels of the high-dimensional raw images. Our theory implies that the linear PCA--ICA cascade can be utilized to identify the true hidden sources of natural image data. This has been demonstrated using video images (Isomura \& Toyoizumi, 2020). In particular, a combination of the proposed theorem with another scheme that effectively removes unpredictable noise from sequential data (i.e., video sequence) is a powerful tool to identify the true hidden sources of real-world data. By this combination, one can obtain data with minimal irrelevant or noise components---which is an ideal condition for the asymptotic linearization theorem---from the original noisy observations. Such a combination enables the reliable and accurate estimations of hidden sources that generate the input sequences, even in the presence of considerable observation noise. The source estimation performance was demonstrated by using sequential visual inputs comprising hand-digits, rotating 3D objects, and natural scenes. These results support the applicability of the proposed theorem to the natural data, and further highlight that the PCA--ICA cascade can extract true or relevant hidden sources when the natural data are sufficiently high-dimensional. Please refer to (Isomura \& Toyoizumi, 2020) for details.

In addition, a few works have cited a preprint version of this paper to justify the application of their linear methods to real-world nonlinear BSS tasks, in the contexts of wireless sensor networks (van der Lee et al., 2019) and single channel source separation (Mika et al., 2020). These works demonstrated that linear methods work well with nonlinear BSS tasks, as predicted by our theorem. The asymptotic linearization theorem is of great importance in justifying the application of these linear methods to real-world nonlinear BSS tasks.

In terms of the interpretability of the outcomes of the PCA--ICA cascade, one may ask if applying the PCA--ICA cascade to natural image patches simply yields less interpretable Gabor-filter-like outputs, which are usually not considered to be certain nonlinear, higher-order features underlying natural images. However, when we applied the PCA--ICA cascade---featuring a separate noise reduction technique that we developed---to natural image data with a high dimensionality reduction rate, we obtained outputs or images that represent features relevant to hidden sources, such as the categories of objects (Isomura \& Toyoizumi, 2020). Based on our observations, three requirements to render the PCA--ICA cascade extract features relevant to hidden sources are considered. First, the dimensionality reduction rate should be high relative to that of the conventional PCA application. Indeed, when the dimensionality reduction rate was low, we observed that the obtained encoders exhibit Gabor-filter-like patterns. Second, the PCA--ICA cascade should be applied not to image patches but to entire images. Image patches contain only a fraction of information about the original hidden sources; thus, it is likely that the original hidden sources cannot be recovered from image patches. Third, PCA should be applied to de-noised data wherein observation noise is removed in advance using a separate scheme, like in our case. When PCA is applied to noisy data, sufficient information for recovering hidden sources cannot be extracted as the major principal components, because PCA preferentially extracts large noise owing to its extra variance. Hence, when the aforementioned requirements are satisfied, the outcome of the PCA--ICA cascade can exhibit a high interpretability, wherein the extracted features are relevant to hidden sources.

As a side note, to utilize the asymptotic linearization theorem to estimate the accuracy of source separation, one cannot use up-sampling techniques to increase the input dimensionality beyond the resolution of the original data or images. This is because these techniques simply yield linear or nonlinear transformations of the original data, which typically do not increase information about true hidden sources. In other words, the resolutions of the original data determine the effective input dimensionality, so that these up-samplings merely render matrix $B$ singular (thereby violating the condition of isotropic $B$).

Furthermore, an interesting possibility is that living organisms might have developed high-dimensional biological sensors to perform nonlinear BSS using a linear biological encoder, because this strategy guarantees robust nonlinear BSS. In particular, this might be the approach taken by the human nervous system using large numbers of sensory cells, such as approximately 100 million rod cells and six million cone cells in the retina, and approximately 16,000 hair cells in the cochlea, to process sensory signals (Kandel et al., 2013).

Neuronal networks in the brain are known to update their synapses by following Hebbian plasticity rules (Hebb, 1949; Malenka \& Bear, 2004), and researchers believe that the Hebbian plasticity plays a key role in representation learning (Dayan \& Abbott, 2001; Gerstner \& Kistler, 2002) and BSS (Brown et al., 2001) in the brain---in particular, the dynamics of neural activity and plasticity in a class of canonical neural networks can be universally characterized in terms of representation learning or BSS from a Bayesian perspective (Isomura \& Friston, 2020). Thus, it follows that major linear BSS algorithms for PCA (Oja, 1982; Oja, 1989) and ICA (Bell \& Sejnowski, 1995; Bell \& Sejnowski, 1997; Amari et al., 1996; Hyvarinen \& Oja, 1997) are formulated as a variant of Hebbian plasticity rules. Moreover, {\it in vitro} neuronal networks learn to separately represent independent hidden sources in (and only in) the presence of Hebbian plasticity (Isomura et al., 2015; Isomura \& Friston, 2018). Nonetheless, the manner in which the brain can possibly solve a nonlinear BSS problem remains unclear, even though it might be a prerequisite for many of its cognitive processes such as visual recognition (DiCarlo et al., 2012). While Oja's subspace rule for PCA (Oja, 1989) and Amari's ICA algorithm (Amari et al., 1996) were used in this paper, these rules can be replaced with more biologically plausible local Hebbian learning rules (Foldiak, 1990; Linsker, 1997; Isomura \& Toyoizumi, 2016; Pehlevan et al., 2017; Isomura \& Toyoizumi, 2018; Leugering \& Pipa, 2018; Isomura \& Toyoizumi, 2019) that require only directly accessible signals to update synapses. A recent work indicated that even a single-layer neural network can perform both PCA and ICA through a local learning rule (Isomura \& Toyoizumi, 2018), implying that even a single-layer network can perform a nonlinear BSS.

In summary, we demonstrated that when the source dimensionality is large and the input dimensionality is sufficiently larger than the source dimensionality, a cascaded setup of linear PCA and ICA can reliably identify the optimal linear encoder to decompose nonlinearly generated sensory inputs into their true hidden sources with increasing accuracy. This is because the higher-dimensional inputs can provide greater evidence about the hidden sources, which removes the possibility of finding spurious solutions and reduces the BSS error. As a corollary of the proposed theorem, Hebbian-like plasticity rules that perform PCA and ICA can reliably update synaptic weights to express the optimal encoding matrix and can consequently identify the true hidden sources in a self-organizing or unsupervised manner. This theoretical justification is potentially useful for designing reliable and explainable artificial intelligence, and for understanding the neuronal mechanisms underlying perceptual inference.


\section{Methods}
\setcounter{equation}{0}

\subsection{Lemma 1}
Suppose $p_s(s)$ is a symmetric distribution, $\tilde{y} \equiv (y_i,y_j)^T \equiv \tilde{A}s$ is a two-dimensional sub-vector of $y = As$, and $\tilde{A} \equiv (A_{i1},\dots,A_{iN_s}; A_{j1},\dots,A_{jN_s})$ is a $2 \times N_s$ sub-matrix of $A$. The expectation of an arbitrary function $F(\tilde{y})$ over $p_s(s)$ is denoted as $\E[F(\tilde{y})]$. Whereas, when $\tilde{y}$ is sampled from a Gaussian distribution $p_\mathcal{N}(\tilde{y}) \equiv \mathcal{N}[0,\tilde{A}\tilde{A}^T]$, the expectation of $F(\tilde{y})$ over $p_\mathcal{N}(\tilde{y})$ is denoted as $\E_\mathcal{N}[F(\tilde{y})] \equiv \int F(\tilde{y}) p_\mathcal{N}(\tilde{y}) d\tilde{y}$. When $F(\tilde{y})$ is order 1, the difference between $\E[F(\tilde{y})]$ and $\E_\mathcal{N}[F(\tilde{y})]$ is upper bounded by order $N_s^{-1}$ as a corollary of the central limit theorem. In particular, $\E[F(\tilde{y})]$ can be expressed as $\E_\mathcal{N}[F(\tilde{y})(1+G(\tilde{y})/N_s)]$, where $G(\tilde{y}) = \mathcal{O}(1)$ is a function that characterizes the deviation of $p(\tilde{y})$ from $p_\mathcal{N}(\tilde{y})$. \\

\noindent{\it Proof.} The characteristic function of $p(\tilde{y})$ is given as $\psi(\tau) \equiv \E[e^{\mathbf{i}\tau^T\tilde{y}}]$ as a function of $\tau \in \mathbb{R}^2$, and its logarithm is denoted as $\Psi(\tau) \equiv \log\psi(\tau)$. The first to fourth order derivatives of $\Psi(\tau)$ with respect to $\tau$ are provided as
\begin{align}
\Psi^{(1)} &= \E[\mathbf{i}\tilde{y}e^{\mathbf{i}\tau^T\tilde{y}}]/\psi,\nonumber\\
\Psi^{(2)} &= \E[(\mathbf{i}\tilde{y})^{\otimes 2}e^{\mathbf{i}\tau^T\tilde{y}}]/\psi - (\Psi^{(1)})^{\otimes 2},\nonumber\\
\Psi^{(3)} &= \E[(\mathbf{i}\tilde{y})^{\otimes 3}e^{\mathbf{i}\tau^T\tilde{y}}]/\psi - \E[(\mathbf{i}\tilde{y})^{\otimes 2}e^{\mathbf{i}\tau^T\tilde{y}}]/\psi \otimes \Psi^{(1)} - \Psi^{(2)} \otimes \Psi^{(1)} - \Psi^{(1)} \otimes \Psi^{(2)}\nonumber\\
&= \E[(\mathbf{i}\tilde{y})^{\otimes 3}e^{\mathbf{i}\tau^T\tilde{y}}]/\psi - 2\Psi^{(2)} \otimes \Psi^{(1)} - (\Psi^{(1)})^{\otimes 3} - \Psi^{(1)} \otimes \Psi^{(2)},\nonumber\\
\Psi^{(4)} &= \E[(\mathbf{i}\tilde{y})^{\otimes 4}e^{\mathbf{i}\tau^T\tilde{y}}]/\psi - \E[(\mathbf{i}\tilde{y})^{\otimes 3}e^{\mathbf{i}\tau^T\tilde{y}}]/\psi \otimes \Psi^{(1)} - 2\Psi^{(3)} \otimes \Psi^{(1)} - 3(\Psi^{(2)})^{\otimes 2} \nonumber\\
&- \Psi^{(2)} \otimes (\Psi^{(1)})^{\otimes 2} - \Psi^{(1)} \otimes \Psi^{(2)} \otimes \Psi^{(1)} - (\Psi^{(1)})^{\otimes 2} \otimes \Psi^{(2)} - \Psi^{(1)} \otimes \Psi^{(3)}.
\end{align}
When $\tau = 0$, they become $\Psi(0) = 0$, $\Psi^{(1)}(0) = 0$, $\Psi^{(2)}(0) = -\E[\tilde{y}^{\otimes 2}]$, $\Psi^{(3)}(0) = 0$, and $\Psi^{(4)}(0) = \E[\tilde{y}^{\otimes 4}] - 3\E[\tilde{y}^{\otimes 2}]^{\otimes 2}$ owing to the symmetric source distribution $p_s(s)$.

They are further computed as $\E[\tilde{y}^{\otimes 2}] = \mathrm{Vec}[\tilde{A}\tilde{A}^T]$ and $\E[\tilde{y}^{\otimes 4}] = \mathrm{Vec}[\E[(\tilde{y}\tilde{y}^T)\otimes(\tilde{y}\tilde{y}^T)]] = 3\mathrm{Vec}[\tilde{A}\tilde{A}^T]^{\otimes 2} + \kappa\mathrm{Vec}[\sum_{k=1}^{N_s} (\tilde{A}_{\bullet k}\tilde{A}_{\bullet k}^T)^{\otimes 2}]$, where $\kappa \equiv \E[s_k^4] - 3$ indicates the kurtosis of the source distribution and $\tilde{A}_{\bullet k}$ denotes the $k$-th column of $\tilde{A}$. Moreover, $\Psi^{(2)}(0) \cdot \tau^{\otimes 2} = -\tau^T \tilde{A}\tilde{A}^T \tau$ and $\Psi^{(4)}(0) \cdot \tau^{\otimes 4} = \kappa\mathrm{Vec}[\sum_{k=1}^{N_s} (\tilde{A}_{\bullet k}\tilde{A}_{\bullet k}^T)^{\otimes 2}] \cdot \tau^{\otimes 4} = \kappa\sum_{k=1}^{N_s} (\tau^T\tilde{A}_{\bullet k})^4 = 3\kappa(\tau^T\tau)^2/N_s + \mathcal{O}(N_s^{-3/2})$ hold. Here, a quartic function of $\tau$, $\sum_{k=1}^{N_s} (\tau^T\tilde{A}_{\bullet k})^4$, is order $N_s^{-1}$ because $\tilde{A}_{\bullet k}$ is sampled from $\mathcal{N}[0,1/N_s]$. Thus, the Taylor expansion of $\Psi(\tau)$ is expressed as
\begin{align}
\Psi(\tau) &= \sum_{n = 0}^\infty \frac{\Psi^{(n)}(0)}{n!} \cdot \tau^{\otimes n} = -\frac{1}{2}\tau^T\tilde{A}\tilde{A}^T\tau + \frac{\kappa}{8N_s}(\tau^T\tau)^2 + \mathcal{O}\left(N_s^{-3/2}\right).
\end{align}
Hence, from $\psi(\tau) = \exp[ -\tau^T\tilde{A}\tilde{A}^T\tau/2 + \kappa(\tau^T\tau)^2/8N_s + \mathcal{O}(N_s^{-3/2}) ] = e^{-\tau^T\tilde{A}\tilde{A}^T\tau/2}( 1 + \kappa(\tau^T\tau)^2/8N_s + \mathcal{O}(N_s^{-3/2}) )$, $p(\tilde{y})$ is computed as follows (the inversion theorem):
\begin{align}
p(\tilde{y}) &= \frac{1}{(2\pi)^2}\int_{\mathbb{R}^2} e^{-\mathbf{i}\tau^T\tilde{y}}\psi(\tau) \lambda(d\tau) \nonumber\\
&= \frac{1}{(2\pi)^2}\int_{\mathbb{R}^2} e^{-\mathbf{i}\tau^T\tilde{y} - \frac{1}{2}\tau^T\tilde{A}\tilde{A}^T\tau} \left( 1 + \frac{\kappa}{8N_s}(\tau^T\tau)^2 + \mathcal{O}\left(N_s^{-3/2}\right) \right) \lambda(d\tau) \nonumber\\
&= p_\mathcal{N}(\tilde{y}) \left( 1 + \frac{G(\tilde{y})}{N_s} \right),
\end{align}
where $\lambda$ denotes the Lebesgue measure. Here, $G(\tilde{y}) = \mathcal{O}(1)$ indicates a function of $\tilde{y}$ that characterizes the difference between $p(\tilde{y})$ and $p_\mathcal{N}(\tilde{y})$. Hence, the expectation over $p(\tilde{y})$ can be rewritten using that over $p_\mathcal{N}(\tilde{y})$:
\begin{align}
\E[F(\tilde{y})] &= \int F(\tilde{y}) p_s(s) ds = \int F(\tilde{y}) p(\tilde{y})d\tilde{y} \nonumber\\
&= \int F(\tilde{y}) p_\mathcal{N}(\tilde{y}) \left( 1 + \frac{G(\tilde{y})}{N_s} \right) d\tilde{y} \nonumber\\
&= \E_\mathcal{N}\left[F(\tilde{y}) \left(1 + \frac{G(\tilde{y})}{N_s}\right) \right].
\end{align}
\hfill $\Box$ \\

\noindent{\it Remark.} For equation \eqref{H}, from the Bussgang theorem (Bussgang, 1952), we obtain
\begin{align}
\E[f(\tilde{y}) \tilde{y}^T] &= \E_\mathcal{N}\left[ f(\tilde{y}) \tilde{y}^T \left( 1 + \frac{G(\tilde{y})}{N_s} \right) \right] \nonumber\\
&= \E_\mathcal{N}\left[ \diag[f'(\tilde{y})] \left( 1 + \frac{G(\tilde{y})}{N_s} \right) + f(\tilde{y}) \frac{G'(\tilde{y})^T}{N_s} \right] \tilde{A}\tilde{A}^T \nonumber\\
&= \left( \diag[\E_\mathcal{N}[f'(\tilde{y})]] \tilde{A} + \mathcal{O}\left(N_s^{-3/2}\right) \right) \tilde{A}^T
\end{align}
as $\tilde{A}$ is order $N_s^{-1/2}$. Thus, we obtain $H = \E[fy^T]A^{+T} = \diag[\E_\mathcal{N}[f']]A + \mathcal{O}(N_s^{-3/2})$.

For equation \eqref{Cov_phi}, because $y_i$ and $y_j$ are only weakly correlated with each other, $(1 + G(\tilde{y})/N_s)$ can be approximated as $(1 + G(\tilde{y})/N_s) \simeq (1 + G(y_i)/N_s)(1 + G(y_j)/N_s)$ as the leading order using newly defined functions $G(y_i)$ and $G(y_j)$. Thus, we obtain
\begin{align}
\Cov[\phi_i, \phi_j] &= \E[(\phi_i - \E[\phi_i])(\phi_j - \E[\phi_j])] \nonumber\\
&\simeq \E_\mathcal{N}\left[ \left\{ (\phi_i - \E[\phi_i])\left( 1 + \frac{G(y_i)}{N_s} \right) \right\} \left\{ (\phi_j - \E[\phi_j])\left( 1 + \frac{G(y_j)}{N_s} \right) \right\} \right] \nonumber\\
&= \Cov_\mathcal{N}[\phi^*_i, \phi^*_j] \nonumber\\
&= \sum_{n=1}^\infty \frac{\E_\mathcal{N}\left[\phi_i^{*(n)}\right] \E_\mathcal{N}\left[\phi_j^{*(n)}\right]}{n!} \Cov_\mathcal{N}[y_i, y_j]^n
\end{align}
for $i \neq j$ as the leading order, where $\phi^*_i \equiv (\phi_i - \E[\phi_i])(1 + G(y_i)/N_s)$ and $\phi^*_j \equiv (\phi_j - \E[\phi_j])(1 + G(y_j)/N_s)$ are modified nonlinear components. The last equality holds according to Lemma 2 (see below), wherein $\E_\mathcal{N}[\phi_i^{*(n)}]$ approximates $\E[\phi_i^{(n)}]$. As a corollary of Lemma 1, the deviation of $\E_\mathcal{N}[\phi_i^{*(n)}] \simeq \E[\phi_i^{(n)}]$ from $\E_\mathcal{N}[\phi_i^{(n)}]$ due to the non-Gaussianity of $y_i$ is order $N_s^{-1}$. Because $\Cov_\mathcal{N}[y_i, y_j]^n = \mathcal{O}(N_s^{-n/2})$, $\E_\mathcal{N}[\phi^{(1)}_i] = \mathcal{O}(N_s^{-1})$, $\E_\mathcal{N}[\phi^{(2)}_i] = \mathcal{O}(N_s^{-1/2})$, and $|\E_\mathcal{N}[\phi^{(n)}_i]| \leq \mathcal{O}(1)$ for $n \geq 3$ hold with a generic odd function $f(\bullet)$ (see Remark in Section 4.2 for details), the difference between $\Cov_\mathcal{N}[\phi^*_i, \phi^*_j]$ and $\Cov_\mathcal{N}[\phi_i, \phi_j]$ is order $N_s^{-5/2}$. Thus, $\Cov[\phi_i, \phi_j]$ can be characterized by computing $\Cov_\mathcal{N}[\phi_i, \phi_j]$ as a proxy up to the negligible deviation; consequently, equation \eqref{Sigma} is obtained.


\subsection{Lemma 2}
Suppose $v$ and $w$ are Gaussian variables with zero mean, $\sigma_v$ and $\sigma_w$ are their standard deviations, and their correlation $c \equiv \Cov[v,w] / \sigma_v \sigma_w$ is smaller than 1. For an arbitrary function $g(\bullet)$,
\begin{align}
\Cov[g(v),g(w)] = \sigma_v \sigma_w \sum_{n = 1}^\infty \E[g^{(n)}(v)]\E[g^{(n)}(w)]\frac{c^n}{n!}.
\end{align}
\\

\noindent{\it Proof.} Because the correlation between $v$ and $w$ is $c$, $w$ can be cast as $w = \sigma_w cv/\sigma_v + \sigma_w\sqrt{1-c^2}\xi$ using a new a zero-mean and unit-variance Gaussian variable $\xi$ that is independent of $v$. When we define the covariance between $g(v)$ and $g(w)$ as $\Psi(c) \equiv \Cov[g(v), g(w)]$, its derivative with respect to $c$ is given as
\begin{align}
\label{Phi_deriv}
\Phi'(c) & = \Cov\left[ g(v), g'\left( \frac{\sigma_w cv}{\sigma_v} + \sigma_w \sqrt{1-c^2}\xi \right) \left( \frac{\sigma_w v}{\sigma_v} - \frac{\sigma_w c\xi}{\sqrt{1-c^2}} \right) \right] \nonumber\\
& = \E\left[ (g(v)-\E[g(v)]) g'\left( \frac{\sigma_w cv}{\sigma_v} + \sigma_w \sqrt{1-c^2}\xi \right) \left( \frac{\sigma_w v}{\sigma_v} - \frac{\sigma_w c\xi}{\sqrt{1-c^2}} \right) \right].
\end{align}
By applying the Bussgang theorem (Bussgang, 1952) with respect to $v$, we obtain
\begin{align}
\label{Phi_deriv1}
& \E\left[ (g(v)-\E[g(v)]) g'\left( \frac{\sigma_w cv}{\sigma_v} + \sigma_w \sqrt{1-c^2}\xi \right) v \right] \nonumber\\
&= \E\left[ g'(v) g'\left( \frac{\sigma_w cv}{\sigma_v} + \sigma_w \sqrt{1-c^2}\xi \right) + (g(v)-\E[g(v)]) g''\left( \frac{\sigma_w cv}{\sigma_v} + \sigma_w \sqrt{1-c^2}\xi \right) \frac{\sigma_w c}{\sigma_v} \right] \sigma_v^2.
\end{align}
Similarly, applying the same theorem with respect to $\xi$ yields
\begin{align}
\label{Phi_deriv2}
& \E\left[ (g(v)-\E[g(v)]) g'\left( \frac{\sigma_w cv}{\sigma_v} + \sigma_w \sqrt{1-c^2}\xi \right) \xi \right] \nonumber\\
&= \E\left[ (g(v)-\E[g(v)]) g''\left( \frac{\sigma_w cv}{\sigma_v} + \sigma_w \sqrt{1-c^2}\xi \right) \sigma_w\sqrt{1-c^2} \right].
\end{align}
As equation \eqref{Phi_deriv} comprises equations \eqref{Phi_deriv1} and \eqref{Phi_deriv2}, $\Phi'(c)$ becomes
\begin{equation}
\phi'(c) = \sigma_v\sigma_w \E\left[ g'(v) g'\left( \frac{\sigma_w cv}{\sigma_v} + \sigma_w \sqrt{1-c^2}\xi \right) \right]
\end{equation}
Hence, for an arbitrary natural number $n$, we obtain
\begin{equation}
\Phi^{(n)}(c) = \sigma_v\sigma_w \E\left[ g^{(n)}(v) g^{(n)}\left( \frac{\sigma_w cv}{\sigma_v} + \sigma_w \sqrt{1-c^2}\xi \right) \right].
\end{equation}
When $c=0$, $\Phi^{(n)}(c) = \sigma_v\sigma_w \E[g^{(n)}(v)]\E[g^{(n)}(\sigma_w \xi)] = \sigma_v\sigma_w \E[g^{(n)}(v)]\E[g^{(n)}(w)]$ holds, where we again regard $\sigma_w \xi = w$. Thus, from the Taylor expansion with respect to $c$, we obtain
\begin{equation}
\Phi(c) = \sum_{n=1}^\infty \Phi^{(n)}(0)\frac{c^n}{n!} = \sigma_v\sigma_w \sum_{n=1}^\infty \E[g^{(n)}(v)] \E[g^{(n)}(w)]\frac{c^n}{n!}.
\end{equation}
\hfill $\Box$ \\

\noindent{\it Remark.} Although equation \eqref{Cov_phi} expresses the expansion with respect to $(\sigma_v \sigma_w c)^n$ instead of $\sigma_v \sigma_w c^n$, their difference is negligible when $\sigma_v$ and $\sigma_w$ are close to 1.

For $\phi(y) = f(y+a) - \E[f] - \E[fy^T] A^{+T} A^+ y$, the coefficients $\E_\mathcal{N}[\phi^{(n)}_i]$ can be computed as follows (up to the fourth order): because applying the Bussgang theorem (Bussgang, 1952) yields $\E_\mathcal{N}[fy^T]A^{+T} A^+ = \diag[\E_\mathcal{N}[f']] = \mathcal{O}(1)$, the difference between $\E[fy^T] A^{+T} A^+$ and $\diag[\E_\mathcal{N}[f']]$ is order $N_s^{-1}$ (due to the non-Gaussian contributions of $y$; c.f., Lemma 1). Thus, from $\phi^{(1)} = \diag[f'] - \E[fy^T] A^{+T} A^+$, we have
\begin{equation}
\E_\mathcal{N}[\phi^{(1)}] = \E_\mathcal{N}\Big[ \diag[f'] - \diag[\E_\mathcal{N}[f']] + \mathcal{O}(N_s^{-1}) \Big] = \mathcal{O}(N_s^{-1}).
\end{equation}
Moreover, for the second to fourth order derivatives, we obtain
\begin{align}
\E_\mathcal{N}[\phi^{(2)}_i] &= \E_\mathcal{N}[f^{(2)}(y_i+a_i)] = \E_\mathcal{N}[f^{(3)}(y_i)]a_i + \mathcal{O}(N_s^{-3/2}) = \mathcal{O}(N_s^{-1/2}), \nonumber\\
\E_\mathcal{N}[\phi^{(3)}_i] &= \E_\mathcal{N}[f^{(3)}(y_i+a_i)] = \E_\mathcal{N}[f^{(3)}(y_i)] + \mathcal{O}(N_s^{-1}) = \mathcal{O}(1), \nonumber\\
\E_\mathcal{N}[\phi^{(4)}_i] &= \E_\mathcal{N}[f^{(4)}(y_i+a_i)] = \mathcal{O}(N_s^{-1/2}).
\end{align}
Here, the Taylor expansion with respect to $a_i$ is applied, where $\E_\mathcal{N}[f^{(2)}(y_i)] = \E_\mathcal{N}[f^{(4)}(y_i)] = 0$ owing to odd function $f(\bullet)$. For $n \geq 5$, $|\E_\mathcal{N}[\phi^{(n)}_i]| \leq \mathcal{O}(1)$ holds. Thus, because $\Cov_\mathcal{N}[y_i, y_j]^n = ((AA^T)^{\odot n})_{ij}$ is order $N_s^{-n/2}$, equation \eqref{Cov_phi} becomes
\begin{align}
\Cov_\mathcal{N}[\phi_i, \phi_j] &= \underbrace{\E_\mathcal{N}[\phi^{(1)}_i]\E_\mathcal{N}[\phi^{(1)}_j](AA^T)_{ij}}_{\mathcal{O}(N_s^{-5/2})} + \frac{1}{2}\E_\mathcal{N}[\phi^{(2)}_i]\E_\mathcal{N}[\phi^{(2)}_j]((AA^T)^{\odot 2})_{ij} \nonumber\\
&+ \frac{1}{6}\E_\mathcal{N}[\phi^{(3)}_i]\E_\mathcal{N}[\phi^{(3)}_j]((AA^T)^{\odot 3})_{ij} + \underbrace{\frac{1}{24}\E_\mathcal{N}[\phi^{(4)}_i]\E_\mathcal{N}[\phi^{(4)}_j]((AA^T)^{\odot 4})_{ij}}_{\mathcal{O}(N_s^{-3})} + \mathcal{O}(N_s^{-5/2}) \nonumber\\
&= \E_\mathcal{N}[f^{(3)}(y_i)]\E_\mathcal{N}[f^{(3)}(y_j)] \left\{ \frac{1}{2}a_i a_j((AA^T)^{\odot 2})_{ij} + \frac{1}{6}((AA^T)^{\odot 3})_{ij} \right\} + \mathcal{O}(N_s^{-5/2}).
\end{align}


\subsection{Lemma 3}
When elements of $A \in \mathbb{R}^{N_f \times N_s}$ are independently sampled from a Gaussian distribution $\mathcal{N}[0,1/N_s]$ and $N_f > N_s^2 \gg 1$, $(AA^T)^{\odot 3} \equiv (AA^T) \odot (AA^T) \odot (AA^T)$ has $N_s$ major eigenmodes---all of which have the eigenvalue of $3N_f/N_s^2$ (as the leading order) with the corresponding eigenvectors that match the directions of $AA^T$---and up to $(N_f-N_s)$ randomly distributed nonzero minor eigenmodes whose eigenvalues are order $\max[1,N_f/N_s^3]$ on average. Thus, the latter are negligibly smaller than the former when $N_f > N_s^2 \gg 1$. \\

\noindent{\it Proof.} Suppose $N_f > N_s^2 \gg 1$. When the $k$-th column vector of $A$ is denoted as $A_{\bullet k} \equiv (A_{1k}, \dots, A_{N_f k})^T \in \mathbb{R}^{N_f}$, $(AA^T)^{\odot 3}$ becomes
\begin{align}
(AA^T)^{\odot 3} & = \sum_{k = 1}^{N_s} \sum_{l = 1}^{N_s} \sum_{m = 1}^{N_s} (A_{\bullet k} \odot A_{\bullet l} \odot A_{\bullet m}) (A_{\bullet k} \odot A_{\bullet l} \odot A_{\bullet m})^T \nonumber\\
& = \sum_{k = 1}^{N_s} (A_{\bullet k}^{\odot 3}) (A_{\bullet k}^{\odot 3})^T + 3 \sum_{k \neq l} (A_{\bullet k}^{\odot 2} \odot A_{\bullet l}) (A_{\bullet k}^{\odot 2} \odot A_{\bullet l})^T \nonumber\\
& \hspace{5mm} + 6\sum_{k < l < m} (A_{\bullet k} \odot A_{\bullet l} \odot A_{\bullet m}) (A_{\bullet k} \odot A_{\bullet l} \odot A_{\bullet m})^T.
\end{align}
Here, $\sum_{k = 1}^{N_s} (A_{\bullet k}^{\odot 3}) (A_{\bullet k}^{\odot 3})^T = (A^{\odot 3}) (A^{\odot 3})^T$ and $\sum_{k \neq l} (A_{\bullet k}^{\odot 2} \odot A_{\bullet l}) (A_{\bullet k}^{\odot 2} \odot A_{\bullet l})^T = \{ (A^{\odot 2}) (A^{\odot 2})^T \} \odot (AA^T) - (A^{\odot 3}) (A^{\odot 3})^T$ hold. Let $QDQ^T$ be the eigenvalue decomposition of $(A^{\odot 2}) (A^{\odot 2})^T$ using a diagonal matrix that arranges eigenvalues in descending order $D \in \mathbb{R}^{N_s \times N_s}$ with the corresponding eigenvectors $Q \in \mathbb{R}^{N_f \times N_s}$. We thus have an expression $\{ (A^{\odot 2}) (A^{\odot 2})^T \} \odot (AA^T) = \sum_{k = 1}^{N_s} \diag(D_{kk}^{1/2} V_{\bullet k}) AA^T \diag(D_{kk}^{1/2} V_{\bullet k})$. Thus, we obtain
\begin{align}
(AA^T)^{\odot 3} &= 3\sum_{k = 1}^{N_s} \diag(D_{kk}^{1/2} V_{\bullet k}) AA^T \diag(D_{kk}^{1/2} V_{\bullet k}) - 2(A^{\odot 3}) (A^{\odot 3})^T \nonumber\\
& \hspace{5mm} + 6 \sum_{k < l < m} (A_{\bullet k} \odot A_{\bullet l} \odot A_{\bullet m}) (A_{\bullet k} \odot A_{\bullet l} \odot A_{\bullet m})^T.
\end{align}
Hence, the first entry in the first term, i.e., $3\diag[D_{11}^{1/2} V_{\bullet 1}] AA^T \diag[D_{11}^{1/2} V_{\bullet 1}] = (3/N_s) \diag[1 + \mathcal{O}(N_s^{-1/2})] AA^T \diag[1 + \mathcal{O}(N_s^{-1/2})]$, corresponds to $N_s$ largest eigenmodes whose eigenvalues are $3N_f/N_s^2$ (as the leading order) with the corresponding eigenvectors that match the directions of $AA^T$. This is because elements of $A$ are independent and identically distributed variables sampled from a Gaussian distribution $\mathcal{N}[0,1/N_s]$. Whereas, the other entries (with $k \geq 2$) in the first term yield $N_s(N_s-1)$ different minor eigenmodes whose eigenvalues are order $N_f/N_s^3$. Then, all the eigenmodes of the second term, $-2(A^{\odot 3}) (A^{\odot 3})^T$, can be expressed using eigenvectors involved in the first term, where all the eigenvalues of $(A^{\odot 3}) (A^{\odot 3})^T$ are order $N_f/N_s^3$. Lastly, the third term involves $N_s(N_s-1)(N_s-2)/6$ terms that are nearly uncorrelated with each other---which are also nearly uncorrelated with the first and second terms---leading to up to $N_s(N_s-1)(N_s-2)/6$ different minor eigenmodes whose eigenvalues are order $N_f/N_s^3$ on average. Intuitively, $A_{\bullet k} \odot A_{\bullet l} \odot A_{\bullet m}$ with $k < l < m$ approximates each eigenmode as elements of $A$ are Gaussian distributed---please ensure that $A_{\bullet k'} \odot A_{\bullet l'} \odot A_{\bullet m'}$ is nearly uncorrelated with $A_{\bullet k} \odot A_{\bullet l} \odot A_{\bullet m}$ if at least one of $(k',l',m')$ is different from $(k,l,m)$. In summary, when $N_s^2 + N_s(N_s-1)(N_s-2)/6 < N_f$, $(AA^T)^{\odot 3}$ has $(N_s(N_s-1) + N_s(N_s-1)(N_s-2)/6)$ randomly distributed nonzero minor eigenmodes whose eigenvalues are of order $N_f/N_s^3$ on average; otherwise, $(AA^T)^{\odot 3}$ has $(N_f-N_s)$ randomly distributed nonzero minor eigenmodes whose eigenvalues are of order 1 on average.
\hfill $\Box$

\noindent{\it Remark.} Because the aforementioned nonzero minor eigenmodes are nearly uncorrelated with eigenmodes of $AA^T$, their contributions are negligible when deriving equation \eqref{Cov_eps2}. It is straightforward to apply this proof to $(AA^T)^{\odot 2}$.


\subsection{First order approximation of major principal components}
The first $N_s$ major eigenmodes of the input covariance approximates the signal covariance comprising the linear projections from every hidden source. We first demonstrate this as the zero-th order approximation by explicitly decomposing the input covariance (equation \eqref{Cov_x}) into the product of orthogonal and diagonal matrices. We define the singular value decomposition of the linear components $BH$ as $BH = U_L S_L V_L^T$ using orthogonal matrices $U_L \in \mathbb{R}^{N_x \times N_s}$ and $V_L \in \mathbb{R}^{N_s \times N_s}$ and a diagonal matrix $S_L \in \mathbb{R}^{N_s \times N_s}$ that arranges singular values in the descending order (note that $N_x > N_s$). Moreover, we define an orthogonal matrix $U_N \in \mathbb{R}^{N_x \times (N_x-N_s)}$ such that it is perpendicular to $U_L$ and that multiplying it by the residual covariance $B\Sigma B^T$ from both sides diagonalizes $B\Sigma B^T$. Thus, $(U_L,U_N)(U_L,U_N)^T = I$ holds and $X_{NN} \equiv U_N^T B\Sigma B^T U_N \in \mathbb{R}^{(N_x-N_s) \times (N_x-N_s)}$ is a diagonal matrix that arranges its diagonal elements in the descending order. Without loss of generality, $B\Sigma B^T$ is decomposed into three matrix factors:
\begin{equation}
\label{BSigmaB}
\begin{split}
B\Sigma B^T = \left( U_L, U_N \right) \left( \begin{array}{cc}
  X_{LL} & X_{NL}^T \\
  X_{NL} & X_{NN}
\end{array} \right) \left( \begin{array}{c}
  U_L^T \\
  U_N^T
\end{array} \right).
\end{split}
\end{equation}
Here, $X_{LL} \equiv U_L^T B\Sigma B^T U_L \in \mathbb{R}^{N_s \times N_s}$ is a symmetric matrix and $X_{NL} \equiv U_N^T B\Sigma B^T U_L \in \mathbb{R}^{(N_x-N_s) \times N_s}$ is a vertically-long rectangular matrix. This decomposition separates $B\Sigma B^T$ into components in the directions of $U_L$ and the rest. From equations \eqref{Cov_x} and \eqref{BSigmaB}, we obtain the key equality that links the eigenvalue decomposition with the decomposition into signal and residual covariances:
\begin{equation}
\label{EVD}
\left( P_M, P_m \right)
\left( \begin{array}{cc}
  \Lambda_M & O \\
  O & \Lambda_m
\end{array} \right)
\left( \begin{array}{c}
  P_M^T \\
  P_m^T
\end{array} \right)
= \left( U_L, U_N \right)
\left( \begin{array}{cc}
  S_L^2 + X_{LL} & X_{NL}^T \\
  X_{NL} & X_{NN}
\end{array} \right)
\left( \begin{array}{c}
  U_L^T \\
  U_N^T
\end{array} \right)
\end{equation}
The left-hand side is the eigenvalue decomposition of $\Cov[x]$, where $\Lambda_M \in \mathbb{R}^{N_s \times N_s}$ and $\Lambda_m \in \mathbb{R}^{(N_x-N_s) \times (N_x-N_s)}$ are diagonal matrices of major and minor eigenvalues, respectively, and $P_M \in \mathbb{R}^{N_x \times N_s}$ and $P_m \in \mathbb{R}^{N_x \times (N_x-N_s)}$ are their corresponding eigenvectors. Whereas, when inequality \eqref{eig_ratio} holds, $S_L^2 + X_{LL}$ is sufficiently close to a diagonal matrix and $X_{NL}$ is negligibly small relative to $S_L^2 + X_{LL}$---thus, the right-hand side can be viewed as the zero-th order approximation of the eigenvalue decomposition of $\Cov[x]$. Owing to the uniqueness of eigenvalue decomposition, we obtain $(P_M, P_m) = (U_L, U_N)$, $\Lambda_M = S_L^2 + \diag[X_{LL}]$, and $\Lambda_m = X_{NN}$ as the leading order.

Next, we compute an error in the first order by considering a small perturbation that diagonalizes the right-hand side of equation \eqref{EVD} more accurately. Suppose $(I, E^T; -E, I) \in \mathbb{R}^{N_x \times N_x}$ as a block matrix that comprises $N_s \times N_s$ and $(N_x-N_s) \times (N_x-N_s)$ identity matrices and a rectangular matrix $E \in \mathbb{R}^{(N_x-N_s) \times N_s}$ with small value elements. As $(I, E^T; -E, I) (I, E^T; -E, I)^T = I + \mathcal{O}(|E|^2)$, it is an orthogonal matrix as the first-order approximation. Because the middle matrix in the right-hand side of equation \eqref{EVD} has been almost diagonalized, a small perturbation by $(I, E^T; -E, I)$ can diagonalize it with higher accuracy. By multiplying $(I, E^T; -E, I)$ and its transpose by the near-diagonal matrix from both sides, we obtain
\begin{equation}
\label{Lambda}
\begin{split}
&\left( \begin{array}{cc}
  I & E^T \\
  -E & I
\end{array} \right)
\left( \begin{array}{cc}
  S_L^2 + X_{LL} & X_{NL}^T \\
  X_{NL} & X_{NN}
\end{array} \right)
\left( \begin{array}{cc}
  I & -E^T \\
  E & I
\end{array} \right) \\
&= \left( \begin{array}{cc}
  S_L^2 + X_{LL} + E^TX_{NL} + X_{NL}^TE & -(S_L^2 + X_{LL}) E^T + X_{NL}^T + E^TX_{NN} \\
  -E(S_L^2 + X_{LL}) + X_{NL} + X_{NN}E & -X_{NL}E^T - EX_{NL}^T + X_{NN}
\end{array} \right)
+ \mathcal{O}(|E|^2)
\end{split}
\end{equation}
up to the first order of $E$. Equation \eqref{Lambda} is diagonalized only when $-E(S_L^2 + X_{LL}) + X_{NL} + X_{NN}E = O$. Thus, we obtain
\begin{equation}
\label{E}
\begin{split}
\mathrm{Vec}[E] = \left\{ I \otimes (S_L^2 + X_{LL}) - X_{NN} \otimes I \right\}^{-1} \mathrm{Vec}[X_{NL}] \approx \left( I \otimes S_L^{-2} \right) \mathrm{Vec}[X_{NL}] \\
\Longleftrightarrow E \approx X_{NL} S_L^{-2} \hspace{50mm}
\end{split}
\end{equation}
as the first-order approximation. Substituting $E = X_{NL} S_L^{-2}$ into equation \eqref{Lambda} yields the first-order approximation of major and minor eigenvalues:
\begin{equation}
\label{eigenvalue_estimator}
\begin{split}
\Lambda_M &= S_L^2 \left( I + S_L^{-2} \mathrm{diag}[X_{LL}] + \mathcal{O}(|E|^2) \right), \\
\Lambda_m &= X_{NN} - 2\mathrm{diag}[X_{NL} S_L^{-2} X_{NL}^T] + \mathcal{O}(|E|^2).
\end{split}
\end{equation}
Moreover, the first-order approximation of their corresponding eigenvectors is provided as:
\begin{equation}
\label{eigenvector_estimator}
\begin{split}
(P_M, P_m) &= (U_L, U_N)
\left( \begin{array}{cc}
  I & -E^T \\
  E & I
\end{array} \right) \\
&= \left( U_L + U_N E, U_N - U_L E^T \right) \\
&= \left( U_L + U_N X_{NL} S_L^{-2}, U_N - U_L S_L^{-2} X_{NL}^T \right) + \mathcal{O}(|E|^2).
\end{split}
\end{equation}
The accuracy of these approximations is empirically validated using numerical simulations (Fig. 2C), wherein an error in approximating $U_L$ using $P_M$ is analytically computed as $\tr[E^TE]/N_s = \tr[S_L^{-2} U_L^T B\Sigma B^T (I - U_LU_L^T) B\Sigma B^T U_L S_L^{-2}]/N_s$ as the leading order. This error is negligibly small relative to the linearization error in equation \eqref{Cov_eps} when inequality \eqref{eig_ratio} holds.


\subsection{Derivation of linearization error covariance}
Equation \eqref{Cov_eps2} is derived as follows: from equation \eqref{H}, $(BH)^+ = \overline{f'}^{-1} (BA)^+ + \mathcal{O}(N_s^{-1})$ holds when $B$ is sufficiently isotropic. Thus, from equations \eqref{Sigma} and \eqref{Cov_eps}, we obtain
\begin{equation}
\Cov[\varepsilon] = \left( \frac{\overline{f^2}}{\overline{f'}^2} - 1 \right) (BA)^+ BB^T (BA)^{+T} + \frac{\overline{f^{(3)}}^2}{2N_s\overline{f'}^2} I
\end{equation}
as the leading order. Here, $aa^T$ and $\Xi$ in equation \eqref{Sigma} are negligibly smaller than the leading order eigenmodes when inequality \eqref{eig_ratio} holds and $B$ is sufficiently isotropic. If we define the left-singular vectors of $A$ as $U_A \in \mathbb{R}^{N_f \times N_s}$, because $A$ has $N_s$ singular values with the leading order term of $(N_f/N_s)^{1/2}$ (Marchenko \& Pastur, 1967), $A = (N_f/N_s)^{1/2}U_A$ holds approximately. Thus, $(BA)^+ = (N_s/N_f)^{1/2}(U_A^TB^TBU_A)^{-1}U_A^TB^T$ holds as the leading order.

Further, as $B$ is isotropic, $U_A^TB^TBU_A$ is sufficiently close to the identity matrix; thus, $(U_A^TB^TBU_A)^{-1} \approx I - U_A^T(B^TB-I)U_A$ provides the first-order approximation. Similarly, $U_A^TB^TBB^TBU_A = I + U_A^T\{2(B^TB-I) + (B^TB-I)^2\}U_A$ is sufficiently close to the identity matrix. Hence,
\begin{equation}
(BA)^+ BB^T (BA)^{+T} \approx \frac{N_s}{N_f} \left\{ I + U_A^T (B^TB-I)^2 U_A \right\}
\end{equation}
holds as the leading order. Thus, using $\Delta \equiv U_A^T (B^TB-I)^2 U_A$, we obtain equation \eqref{Cov_eps2}.


\subsection{Hebbian-like learning rules}
Oja's subspace rule for PCA (Oja, 1989) is defined as
\begin{equation}
\label{Oja}
\dot{W}_{\rm PCA} \propto \E\left[ u_{\rm PCA} (x - \E[x] - W_{\rm PCA}^T u_{\rm PCA})^T \right],
\end{equation}
where $W_{\rm PCA} \in \mathbb{R}^{N_s \times N_x}$ is a synaptic weight matrix, $u_{\rm PCA} \equiv W_{\rm PCA}(x - \E[x]) \in \mathbb{R}^{N_s}$ is a vector of encoders, and the dot over $W_{\rm PCA}$ denotes a temporal derivative. This rule can extract a subspace spanned by the first $N_s$ principal components (i.e., $W_{\rm PCA} \to \Omega_M P_M^T$) in a manner equal to PCA via eigenvalue decomposition. Although Oja's subspace rule does not have a cost function, a gradient descent rule for PCA---based on the least mean squared error---is known (Xu, 1993), whose cost function is given as $\E[ | x - \E[x] - W_{\rm PCA}^T u_{\rm PCA} |^2 ]$. Indeed, this equals the maximum likelihood estimation of $x - \E[x]$ using a lower-dimensional linear encoder $u_{\rm PCA}$ under the assumption that the loss follows a unit Gaussian distribution. The gradient descent on this cost function is equal to Oja's subspace rule up to an additional term that does not essentially change the behavior of the algorithm.

For BSS of $u_{\rm PCA}$, Amari's ICA algorithm is considered (Amari et al., 1996). Using encoders $\tilde{u} \equiv W_{\rm ICA} u_{\rm PCA} \in \mathbb{R}^{N_s}$ with synaptic weight matrix $W_{\rm ICA} \in \mathbb{R}^{N_s \times N_s}$, the ICA cost function is defined as the Kullback--Leibler divergence between the actual distribution of $\tilde{u}$, $p(\tilde{u})$, and its prior belief $p_0(\tilde{u}) \equiv \prod_i p_0(\tilde{u}_i)$, $\mathcal{D}_{\rm KL} \equiv \mathcal{D}_{\rm KL}[p(\tilde{u}) || p_0(\tilde{u})] \equiv \E[ \log p(\tilde{u}) - \log p_0(\tilde{u}) ]$. The natural gradient of $\mathcal{D}_{\rm KL}$ yields Amari's ICA algorithm,
\begin{equation}
\dot{W}_{\rm ICA} \propto -\frac{\partial \mathcal{D}_{\rm KL}}{\partial W_{\rm ICA}}W_{\rm ICA}^TW_{\rm ICA} = (I - \E\left[ g(\tilde{u}) \tilde{u}^T \right]) W_{\rm ICA},
\end{equation}
where $g(\tilde{u}) \equiv -\mathrm{d}\log p_0(\tilde{u})/\mathrm{d}\tilde{u}$ is a nonlinear activation function.


\section*{Data Availability}
All relevant data are within the paper. The MATLAB scripts are available at \url{https://github.com/takuyaisomura/asymptotic_linearization}.

\section*{Acknowledgements}
This work was supported by RIKEN Center for Brain Science (T.I. and T.T.), Brain/MINDS from AMED under Grant Number JP20dm020700 (T.T.), and JSPS KAKENHI Grant Number JP18H05432 (T.T.). The funders had no role in the study design, data collection and analysis, decision to publish, or preparation of the manuscript.


\section*{References}
\leftskip = 20pt
\parindent = -20pt

\hspace{-24pt} Amari, S.I., Chen, T.P. \& Cichocki, A. (1997) Stability analysis of learning algorithms for blind source separation. Neural Netw 10(8):1345-1351.

Amari, S.I., Cichocki, A. \& Yang, H.H. (1996) A new learning algorithm for blind signal separation. Adv Neural Inf Proc Sys 8:757-763.

Arora, S. \& Risteski, A. (2017) Provable benefits of representation learning. arXiv:1706.04601.

Baldi, P. \& Hornik, K. (1989) Neural networks and principal component analysis: Learning from examples without local minima. Neural Netw 2(1):53-58.

Barron, A.R. (1993) Universal approximation bounds for superpositions of a sigmoidal function. IEEE Trans Info Theory 39(3):930-945.

Bell, A.J. \& Sejnowski, T.J. (1995) An information-maximization approach to blind separation and blind deconvolution. Neural Comput 7(6):1129-1159.

Bell, A.J. \& Sejnowski, T.J. (1997) The ``independent components'' of natural scenes are edge filters. Vision Res 37(23):3327-3338.

Brown, G.D., Yamada, S. \& Sejnowski, T.J. (2001) Independent component analysis at the neural cocktail party. Trends Neurosci 24(1):54-63.

Bussgang, J.J. (1952) Cross-correlation functions of amplitude-distorted Gaussian signals. Res Lab Elec, Mas Inst Technol, Cambridge MA, Tech Rep 216.

Calhoun, V.D., Liu, J. \& Adali, T. (2009) A review of group ICA for fMRI data and ICA for joint inference of imaging, genetic, and ERP data. Neuroimage 45(1):S163-S172.

Chen, T., Hua, Y. \& Yan, W.Y. (1998) Global convergence of Oja's subspace algorithm for principal component extraction. IEEE Trans Neural Netw 9(1):58-67.

Chandler, D.M. \& Field, D.J. (2007) Estimates of the information content and dimensionality of natural scenes from proximity distributions. J Opt Soc Am A 24(4):922-941.

Cichocki, A., Zdunek, R., Phan, A.H. \& Amari, S.I. (2009) Nonnegative Matrix and Tensor Factorizations: Applications to Exploratory Multi-way Data Analysis and Blind Source Separation. (John Wiley \& Sons).

Comon, P. (1994) Independent component analysis, a new concept?. Sig Process 36(3):287-314.

Comon, P. \& Jutten, C. (2010) Handbook of Blind Source Separation: Independent Component Analysis and Applications. (Academic Press, Orlando, FL, USA).

Cybenko, G. (1989) Approximation by superpositions of a sigmoidal function. Math Control Signals Syst 2(4):303-314.

Dahl, G.E., Yu, D., Deng, L. \& Acero, A. (2012) Context-dependent pre-trained deep neural networks for large-vocabulary speech recognition. IEEE Trans Audio Speech Lang Proc 20(1):30-42.

Dayan, P. \& Abbott, L.F. (2001) Theoretical neuroscience: computational and mathematical modeling of neural systems (MIT Press, London).

Dayan, P., Hinton, G.E., Neal, R.M. \& Zemel, R.S. (1995) The Helmholtz machine. Neural Comput 7(5):889-904.

DiCarlo, J.J., Zoccolan, D. \& Rust, N.C. (2012) How does the brain solve visual object recognition? Neuron 73(3):415-434.

Dinh, L., Krueger, D. \& Bengio, Y. (2014) NICE: Non-linear independent components estimation. arXiv:1410.8516.

Erdogan, A.T. (2007) Globally convergent deflationary instantaneous blind source separation algorithm for digital communication signals. IEEE Trans Signal Process 55(5):2182-2192.

Erdogan, A.T. (2009) On the convergence of ICA algorithms with symmetric orthogonalization. IEEE Trans Signal Process 57(6):2209-2221.

F\"oldi\'ak, P. (1990) Forming sparse representations by local anti-Hebbian learning. Biol Cybern 64(2):165-170.

Friston, K. (2008) Hierarchical models in the brain. PLoS Comput. Biol 4:e1000211.

Friston, K., Trujillo-Barreto, N. \& Daunizeau, J. (2008) DEM: A variational treatment of dynamic systems. NeuroImage 41(3):849-885.

Gerstner, W. \& Kistler, W.M. (2002) Spiking Neuron Models: Single Neurons, Populations, Plasticity, (Cambridge University Press, Cambridge).

Goodfellow, I., Bengio, Y., \& Courville, A. (2016) Deep learning. (Cambridge: MIT press).

Griffiths, D.J. (2005) Introduction to quantum mechanics. 2nd ed. (Pearson Prentice Hall).

Hebb, D.O. (1949) The Organization of Behavior: A Neuropsychological Theory (Wiley, New York).

Hinton, G.E. \& Salakhutdinov, R.R. (2006) Reducing the dimensionality of data with neural networks. Science 313(5786):504-507.

Hinton, G.E., Srivastava, N., Krizhevsky, A., Sutskever, I. \& Salakhutdinov, R.R. (2012) Improving neural networks by preventing co-adaptation of feature detectors. arXiv:1207.0580.

Hornik, K., Stinchcombe, M. \& White, H. (1989) Multilayer feedforward networks are universal approximators. Neural Netw, 2(5):359-366.

Hyv\"arinen, A. \& Morioka, H. (2016) Unsupervised feature extraction by time-contrastive learning and nonlinear ica. Adv Neural Info Proc Sys 3765-3773.

Hyv\"arinen, A.J. \& Morioka, H. (2017) Nonlinear ICA of temporally dependent stationary sources. Proc Machine Learn Res.

Hyv\"arinen, A. \& Oja, E. (1997) A fast fixed-point algorithm for independent component analysis. Neural Comput 9(7):1483-1492.

Hyv\"arinen, A. \& Pajunen, P. (1999) Nonlinear independent component analysis: Existence and uniqueness results. Neural Netw 12(3):429-439.

Isomura, T. \& Friston, K. (2018). In vitro neural networks minimise variational free energy. Sci Rep 8, 16926.

Isomura, T. \& Friston, K. (2020). Reverse engineering neural networks to characterise their cost functions. Neural Comput 32(11):2085-2121.

Isomura, T., Kotani, K. \& Jimbo, Y. (2015) Cultured cortical neurons can perform blind source separation according to the free-energy principle. PLoS Comput Biol 11(12):e1004643.

Isomura, T. \& Toyoizumi, T. (2016) A local learning rule for independent component analysis. Sci Rep 6:28073.

Isomura, T. \& Toyoizumi, T. (2018) Error-gated Hebbian rule: a local learning rule for principal and independent component analysis. Sci Rep 8:1835.

Isomura, T. \& Toyoizumi, T. (2019) Multi-context blind source separation by error-gated Hebbian rule. Sci Rep 9:7127.

Isomura, T. \& Toyoizumi, T. (2020) Dimensionality reduction to maximize prediction generalization capability. arXiv:2003.00470.

Jolliffe, I.T. (2002) Principal Component Analysis (2nd ed.). (Springer).

Jutten, C. \& Karhunen, J. (2004) Advances in blind source separation (BSS) and independent component analysis (ICA) for nonlinear mixtures. Int J Neural Syst 14(5):267-292.

Kandel, E.R., Schwartz, J.H., Jessell, T.M., Siegelbaum, S.A. \& Hudspeth, A.J. (2013) Principles of Neural Science 5th edn. (McGraw-Hill, New York).

Karhunen, J. (2001). Nonlinear independent component analysis. Independent Component Analysis: Principles and Practice, eds Roberts, S. \& Everson, R. (Cambridge University Press, Cambridge) pp 113-134.

Kawaguchi, K. (2016) Deep learning without poor local minima. Adv Neural Inf Proc Sys 29:586-594.

Khemakhem, I., Kingma, D., Monti, R. \& Hyvarinen, A. (2020). Variational autoencoders and nonlinear ICA: A unifying framework. In International Conference on Artificial Intelligence and Statistics 2207-2217.

Kingma, D.P. \& Welling, M. (2013) Auto-encoding variational bayes. arXiv:1312.6114.

Lappalainen, H. \& Honkela, A. (2000) Bayesian non-linear independent component analysis by multi-layer perceptrons. Advances in Independent Component Analysis (Springer, London), pp 93-121.

Leugering, J. \& Pipa, G. (2018) A unifying framework of synaptic and intrinsic plasticity in neural populations. Neural Comput 30(4):945-986.

Linsker, R. (1997) A local learning rule that enables information maximization for arbitrary input distributions. Neural Comput 9(8):1661-1665.

Lu, H. \& Kawaguchi, K. (2017) Depth creates no bad local minima. arXiv:1702.08580.

Malenka, R.C. \& Bear, M.F. (2004) LTP and LTD: an embarrassment of riches. Neuron 44(1):5-21.

Marchenko, V.A. \& Pastur, L.A. (1967) Distribution of eigenvalues for some sets of random matrices. Mat Sbornik 114:507-536.

Mika, D., Budzik, G., \& J\'ozwik, J. (2020) Single Channel Source Separation with ICA-Based Time-Frequency Decomposition. Sensors, 20(7), 2019.

Nguyen, Q. \& Hein, M. (2017) The loss surface of deep and wide neural networks. arXiv:1704.08045.

Oja, E. (1982) Simplified neuron model as a principal component analyzer. J Math Biol 15(3):267-273.

Oja, E. (1989) Neural networks, principal components, and subspaces. Int J Neural Syst 1(10):61-68.

Oja, E. \& Yuan, Z. (2006) The FastICA algorithm revisited: Convergence analysis. IEEE Trans Neural Netw 17(6): 1370-1381.

Papadias, C.B. (2000) Globally convergent blind source separation based on a multiuser kurtosis maximization criterion. IEEE Trans Signal Process 48(12):3508-3519.

Pearson, K. (1901) On lines and planes of closest fit to systems of points in space. Philos Mag 2(11):559-572.

Pehlevan, C., Mohan, S. \& Chklovskii, D.B. (2017) Blind nonnegative source separation using biological neural networks. Neural Comput 29(11):2925-2954.

Rahimi, A. \& Recht, B. (2008a) Uniform approximation of functions with random bases. In Proceedings of the 46th Annual Allerton Conference on Communication, Control, and Computing:555-561.

Rahimi, A. \& Recht, B. (2008b) Weighted sums of random kitchen sinks: Replacing minimization with randomization in learning. Adv Neural Info Process Sys 21:1313-1320.

Sanger, T.D. (1989) Optimal unsupervised learning in a single-layer linear feedforward neural network. Neural Netw 2(6):459-473.

Toyoizumi, T. \& Abbott, L.F. (2011) Beyond the edge of chaos: Amplification and temporal integration by recurrent networks in the chaotic regime. Phys Rev E 84(5):051908.

van der Lee, T., Exarchakos, G., \& de Groot, S. H. (2019) In-network Hebbian plasticity for wireless sensor networks. In International Conference on Internet and Distributed Computing Systems (pp. 79-88). Springer, Cham.

Wan, L., Zeiler, M., Zhang, S., LeCun, Y. \& Fergus, R. (2013) Regularization of neural networks using DropConnect. In Proceedings of Machine Learning Research (28)3:1058-1066.

Wentzell, P. D., Andrews, D. T., Hamilton, D. C., Faber, K. \& Kowalski, B. R. (1997) Maximum likelihood principal component analysis. J Chemom 11(4):339-366.

Xu, L. (1993) Least mean square error reconstruction principle for self-organizing neural-nets. Neural Netw 6(5):627-648.


\end{document}